\documentclass[10pt,twocolumn,letterpaper]{article}
\usepackage{booktabs}
\usepackage{multirow}
\usepackage{graphicx}
\usepackage{makecell}
\usepackage{float}
\usepackage{colortbl}
\usepackage[table]{xcolor}

\usepackage{cvpr}

\definecolor{cvprblue}{rgb}{0.21,0.49,0.74}
\usepackage[pagebackref,breaklinks,colorlinks,allcolors=cvprblue]{hyperref}

\title{SAIDO: Generalizable Detection of AI-Generated Images via Scene-Aware and Importance-Guided Dynamic Optimization in Continual Learning}

\author{
Yongkang Hu$^{1,\dagger}$\quad
Yu Cheng$^{1,2,\dagger}$\quad
Yushuo Zhang$^{1}$ \quad
Yuan Xie$^{1,2}$\quad
Zhaoxia Yin$^{1,*}$ \\
\\[-6pt]
$^{1}$East China Normal University \\
$^{2}$Shanghai Innovation Institute \\
{\tt\small 10242140409@stu.ecnu.edu.cn, yucheng@sii.edu.cn, zxyin@cee.ecnu.edu.cn} \\
\\[-8pt]
}

\begin{document}
\maketitle
\begin{abstract}
The widespread misuse of image generation technologies has raised security concerns, driving the development of AI-generated image detection methods. However, generalization has become a key challenge and open problem: existing approaches struggle to adapt to emerging generative methods and content types in real-world scenarios. To address this issue, we propose a Scene-Aware and Importance-Guided Dynamic Optimization detection framework with continual learning (SAIDO). Specifically, we design Scene-Awareness-Based Expert Module (SAEM) that dynamically identifies and incorporates new scenes using VLLMs. For each scene, independent expert modules are dynamically allocated, enabling the framework to capture scene-specific forgery features better and enhance cross-scene generalization. To mitigate catastrophic forgetting when learning from multiple image generative methods, we introduce Importance-Guided Dynamic Optimization Mechanism (IDOM), which optimizes each neuron through an importance-guided gradient projection strategy, thereby achieving an effective balance between model plasticity and stability. Extensive experiments on continual learning tasks demonstrate that our method outperforms the current SOTA method in both stability and plasticity, achieving 44.22\% and 40.57\% relative reductions in average detection error rate and forgetting rate, respectively. On open-world datasets, it improves the average detection accuracy by 9.47\% compared to the current SOTA method.
\end{abstract}    
\section{Introduction}
\label{sec:intro}

\begin{figure}[htbp]
    \centering
    \includegraphics[width=\linewidth]{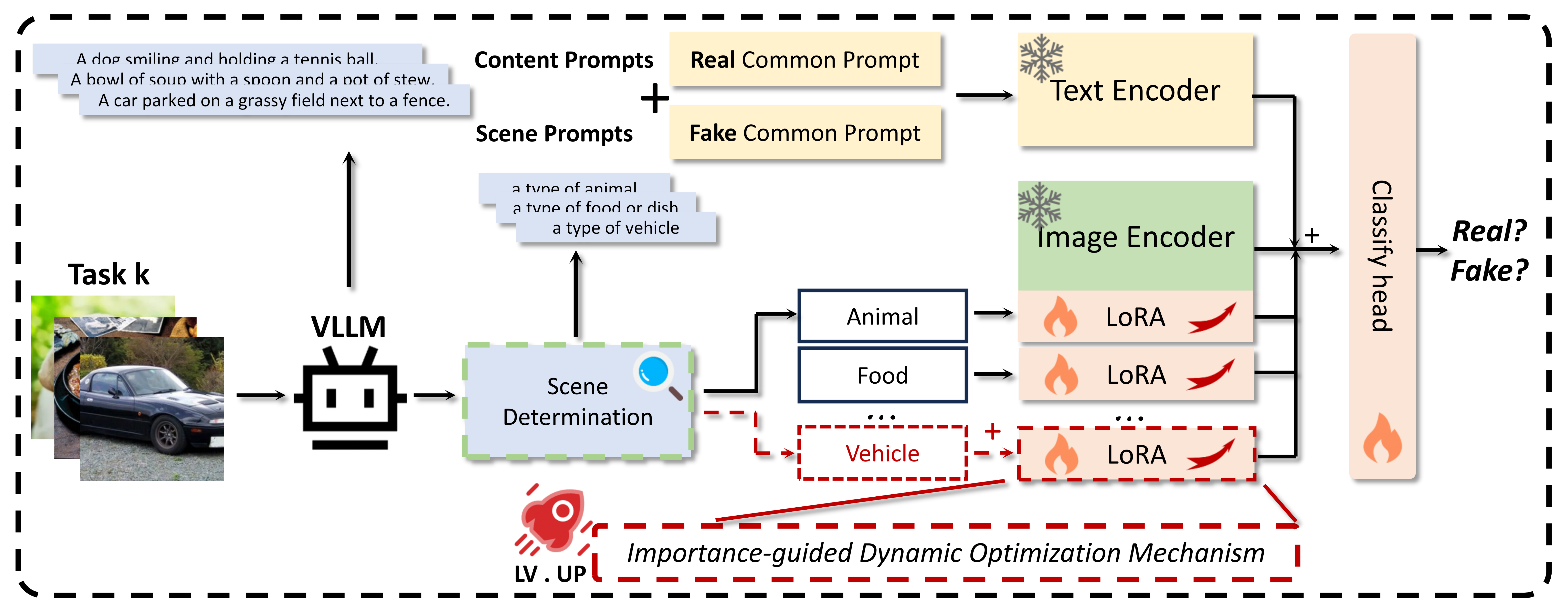} 
    \caption{The pipeline of the proposed framework}
    \label{fig:pipline}
\end{figure}

In recent years, AIGC technologies have rapidly advanced, especially image-based generation methods \cite{wang2025gifdl,2024migc}, which have been widely applied in daily life. However, the misuse of such technologies for malicious purposes, such as fake news dissemination, poses serious security threats, highlighting the urgent need for effective detection methods.

Existing detection methods primarily rely on data-driven discriminative mechanisms, where models learn the differences between real and AI-generated images in terms of frequency distribution \cite{2025frequency}, texture structure \cite{Gram}, and semantic consistency \cite{2025c2p}, thereby establishing decision boundaries in high-dimensional feature space. These methods have achieved promising results under closed-set conditions. However, with the rapid evolution of generative models such as ProGAN \cite{karras2018progressive}, Stable Diffusion \cite{rombach2022high} and other types, the distributional gap between generated domains has continued to widen \cite{domainshift}. Existing detection models often suffer from severe performance degradation due to catastrophic forgetting when adapting to new generative models and contents.

Current detection methods with continual learning provide an effective solution to the problem mentioned above. Pan et al. \cite{DFIL} proposed DFIL, which uses both center and hard samples for replay to alleviate catastrophic forgetting, achieving promising initial results. Gao et al. \cite{DynamicSwin} sought to address this problem by building independent detectors for each new generative method within an incremental learning framework. Tang et al. \cite{tifs} further introduced a content-agnostic adapter-based continual detection framework, which achieved improved performance over prior methods. However, these methods struggle to meet the requirements of real-world detection scenarios, where storing and replaying previously seen samples are often infeasible, and exhibit limited extensibility in efficiently adapting to new generative methods, resulting in limited generalization and stability when dealing with diverse scene domains and multiple generative methods.

To address these limitations, we propose the SAIDO, as shown in \cref{fig:pipline},  which aims to achieve efficient adaptation and continual learning across multiple generative models and diverse scene distributions in real-world environments. Specifically, Vision Language Large Models (VLLMs) are employed first to extract scene features and dynamically identify both existing and newly emerging scenes. Then, for each scene, an independent lightweight LoRA expert module is dynamically allocated. To update these LoRA modules, the IDOM is introduced, which precisely adjusts neuron update directions based on their importance in historical tasks, enabling fine-grained control at the neuron level. Concretely, the method measures neuron importance to distinguish core and non-core neurons, applying importance-guided gradient projection strategy to core neurons while allowing non-core neurons to update freely, thus achieving a ``stability–plasticity” dynamic balance at the neuron level. Experimental results show that SAIDO exhibits outstanding detection performance and continual learning capability across diverse generative models and complex scenarios, providing an efficient and extensible solution to open-world AI-generated image detection. The contributions of this paper are summarized as follows:

\begin{itemize}[label=$\bullet$, leftmargin=*]
    \item The proposed SAEM utilizes VLLMs to extract scene representations for dynamically distinguishing and updating scenes. For each scene, an independent LoRA expert model is constructed, enabling efficient adaptation and strong generalization across diverse domains.
    \item The IDOM serves as an effective strategy for fine-tuning LoRA expert models. It measures neuron importance and applies differentiated parameter update constraints, thereby achieving an efficient balance between plasticity and stability.
    \item Extensive experiments demonstrate that the SAIDO framework surpasses current SOTA methods in both stability and plasticity, achieving relative reductions of 44.22\% in average detection error rate and 40.57\% in forgetting rate, while further improving the average detection accuracy on open-world datasets by 9.47\%.
\end{itemize}

\section{Related Work}
\subsection{AI-generated Image Detection}
AI-generated image detection \cite{yu2024semgir,lu2025liteupdate} aims to determine whether an image is produced by a generative model, addressing the risks of AI image misuse. Early advances in deep neural networks significantly improved the detection of forgery traces \cite{Cnndetection}.
Later works explored universal artifacts in both spatial \cite{Gram,2022kongjian} and frequency domains \cite{2021pinyu,2024pinyu}.
Some studies further analyzed the feature discrepancies between GANs and diffusion models \cite{2023gan-df}, and investigated either high-dimensional, content-agnostic representations \cite{2023clipdetection,2023LGrad} or fine-grained pixel-level structures \cite{2024NPR}. Other works further examined the link between universal forgery clues and high-level semantics \cite{AIDE,2025c2p}.

However, with the rapid evolution of image generative models, existing AI-generated image detection methods struggle to adapt to newly emerging generative methods. Retraining the detectors on new data often leads to catastrophic forgetting, resulting in a severe degradation in detection performance. Therefore, the detection methods with continual learning offer a more promising solution for AI-generated image detection.

\subsection{AI-generated Image Detection with Continual Learning}
Continual learning has been extensively studied, with various paradigms proposed, such as data replay \cite{2021reply}, parameter regularization \cite{2017ewc,2017lwf}, and gradient projection approaches \cite{2021gradient,2024gp,2025gradient}. However, research on AI-generated image detection methods with continual learning is limited. CoReD \cite{2021cored} mitigates catastrophic forgetting through a distillation-based loss, while DFIL \cite{DFIL} combines multi-perspective knowledge distillation with center and hard sample replay strategies to address this issue. S-Prompts \cite{sprompts} achieves knowledge transfer across different tasks through learnable soft prompts. SUR-LID \cite{stack_cvpr} aims to maintain the global distribution of old task features through a sparse and uniform replay strategy. DynamicSwin \cite{DynamicSwin} employs independently updated multi-adapter structures to alleviate catastrophic forgetting and enhance generalization, whereas Tang et al. \cite{tifs} further tackle this challenge by introducing content-agnostic adapters. Rego \cite{RegO} mitigates catastrophic forgetting through gradient projection theory but relies heavily on a threshold-based classification scheme, which limits precise neuron-level regulation.

Nevertheless, existing methods with continual learning struggle to achieve an effective balance between model plasticity and stability, and most rely on replaying historical data to mitigate catastrophic forgetting, which prevents them from meeting the demands of real-world scenarios.
\section{Methodology}

\begin{figure*}[t]
  \centering
  \includegraphics[width=0.95\textwidth]{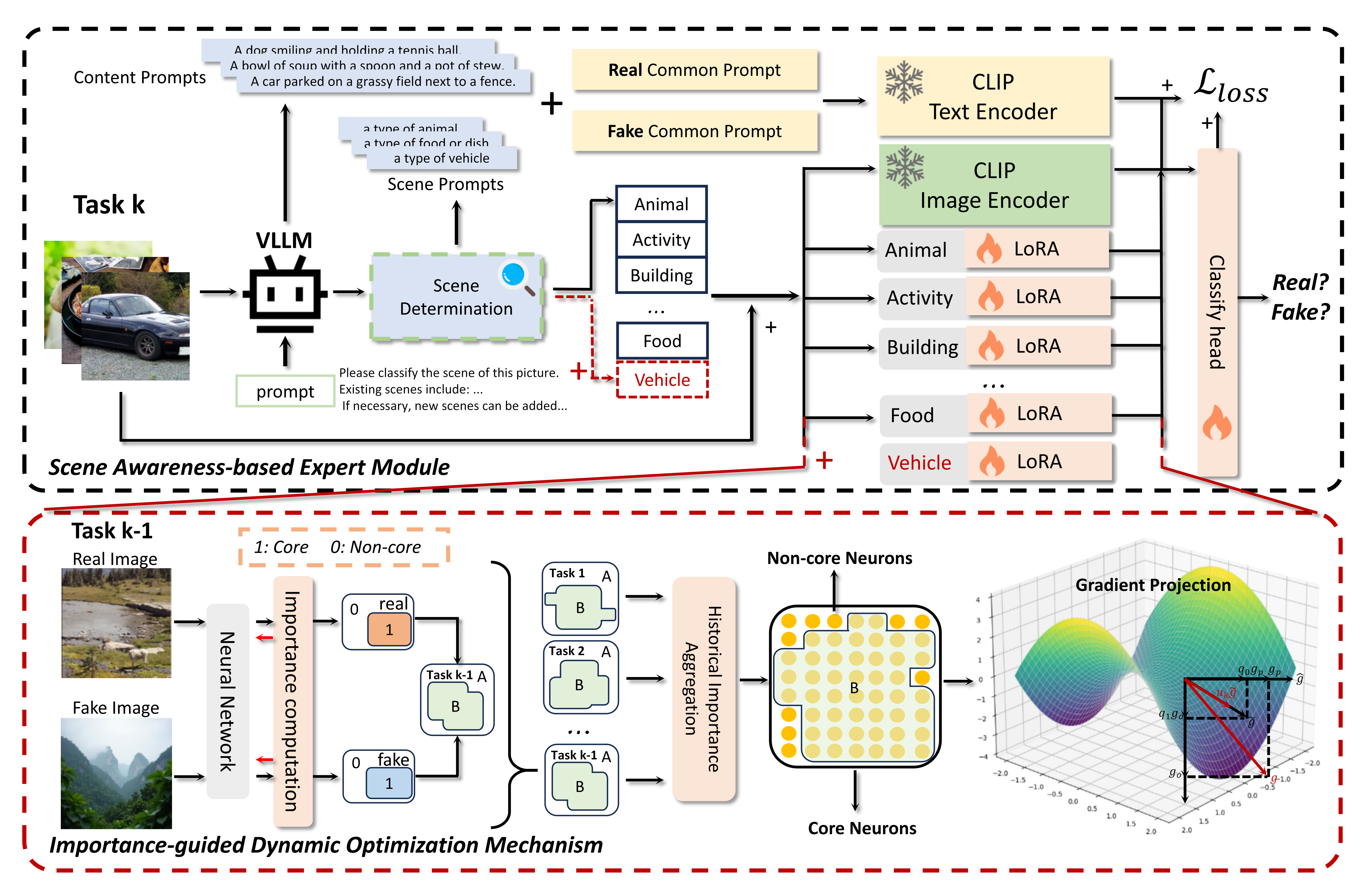}
  \caption{Overall framework of the proposed method.}
  \label{fig:overall}
\end{figure*}

\subsection{Overall Framework}
\begin{table}[t]
\centering
\caption{Notations.}
\label{tab:notations}
\resizebox{\columnwidth}{!}{%
\begin{tabular}{cc}
\toprule
\textbf{Notation} & \textbf{Description} \\
\midrule
$x$ & Input image \\
$l$ & textual prompts \\
$S$ & Scene set \\
$F$ & Fisher information matrix \\
$I_k$ & Neuron importance matrix of the $k$-th task \\
$M_k$ & Neuron importance mask matrix \\
$Q_\alpha$ & The $\alpha$-quantile function \\
$\odot$ & Hadamard product \\
$g$ & original gradient \\
$g_A$ & Optimized gradient of core neurons \\
$g_B$ & Optimized gradient of non-core neurons \\
\bottomrule
\end{tabular}%
} 
\end{table}

To address the challenges of adapting to diverse real-world scenes and continuously emerging generative methods, we propose a Scene-Aware and Importance-guided Dynamic Optimization (SAIDO) detection framework with continual learning, which consists of two key components: the Scene-Aware Based Expert Module (SAEM) and Importance-guided Dynamic Optimization Mechanism (IDOM), as shown in \cref{fig:overall}. In SAEM, VLLMs are utilized to extract scene features and classify images into corresponding scene categories. For each identified scene, an independent LoRA expert model is dynamically allocated. When a new scene emerges, the scene library is updated accordingly, and a new LoRA expert is assigned to it. Each expert model is dedicated to detecting forgery traces within its specific scene. The IDOM is responsible for gradient updates of the LoRA models. It evaluates the importance of each neuron with respect to real and generated images, which is dynamically adjusted based on historical tasks, categorizing them into “core” and “non-core” neurons. Core neurons are updated through importance-guided gradient projection strategy, while non-core neurons are allowed to update freely.

\subsection{Scene-Awareness-Based Expert Module}
In real-world image forgery detection tasks, models must handle complex image content originating from multiple scene domains, where scene-domain shifts in content may lead to performance degradation. Moreover, in open-world settings, the diversity and dynamically evolving nature of image content make manually defined or fixed classifiers insufficient to cover all potential scenes.

To address this, we propose the SAEM, which employs a CLIP model as the backbone and dynamically allocates independent lightweight LoRA expert modules to different scenes. Each expert module learns scene-specific features and forgery traces, thereby maintaining strong generalization and adaptability across diverse content domains. Specifically, SAEM uses a VLLM for scene awareness, dynamically identifying the scene category of each input image and updating the scene library when a new scene is detected. Meanwhile, the VLLM generates a textual description $l_{content}$ for the image’s scene information.
\begin{figure}[t]
    \centering
    \includegraphics[width=\linewidth]{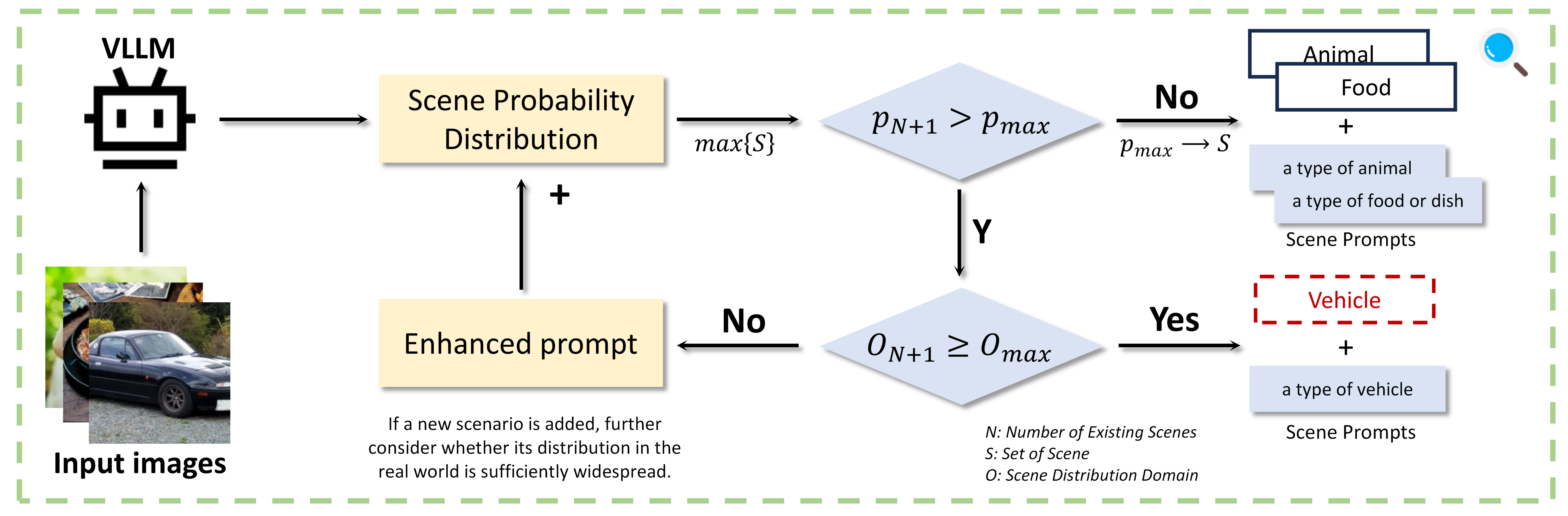} 
    \caption{The process of scene determination. Through a multi-level discrimination mechanism, VLLMs are utilized to achieve scene classification and dynamic expansion to new scenes.}
    \label{fig:scene determination}
\end{figure}

\begin{equation}
    p = \text{VLLM}(x) 
    \label{eq:VLM}
\end{equation}

As shown in \cref{fig:scene determination}, after processing the input image with the VLLM for scene awareness, the confidence distribution over all scenes is represented as $p = [p_1, p_2, \dots, p_N, p_{N+1}] \in \mathbb{R}^{N+1}$, where $N$ denotes the number of currently recognized scenes, and $p_{N+1}$ indicates the confidence of a newly emerging scene. The VLLM assigns the scene label of the image based on the category with the highest confidence and automatically produces the corresponding scene prompt $l_{scene}$. We thus set common prompts $l_{common}$ such as ``real" and ``fake" to distinguish between real and fake features in the high-level semantic space. To enhance model guidance in the semantic space, we introduce Scene-Aware Prompts (SAPs), which integrate three complementary components: the content description $l_{content}$, the scene prompt $l_{scene}$, and the common prompts $l_{common}$.
\begin{equation}
    l_{SAP} = [l_{content}; l_{scene}; l_{common}]
\end{equation}

For images classified into known categories, the corresponding LoRA expert module serves as the detector. When a new scene is identified, a new scene label is automatically created, and a LoRA expert model
${\phi}_{N+1}$ is assigned. In this Module, the LoRA expert modules for all scenes are denoted as $\hat{\phi} \in \{\phi_1, \phi_2, \dots, \phi_N\}$, and the CLIP backbone network is represented as $\phi$. By combining the backbone network, the scene-specific LoRA modules, and the shared classification head, we obtain the final complete model structure: $f_{\phi + \hat{\phi}} \in \{f_{\phi + \phi_1}, f_{\phi + \phi_2}, \dots, f_{\phi + \phi_N}\}$.

Once the image scene is determined by the VLLM, both the scene category of the input image $x$ and its corresponding LoRA expert module are established, enabling the image to be processed by the appropriate scene-specific detector:
\begin{equation}
\phi_n \leftarrow \max [p_1, p_2, \ldots, p_N, p_{N+1}]
\end{equation}

Subsequently, the image is fed into our detection model, processed by the corresponding scene-specific LoRA, and finally passed through the shared classification head to produce the final output, expressed as:
\begin{equation}
\hat{y} = f_{\phi + \phi_n}(\boldsymbol{x})
\end{equation}

\noindent
where \( \hat{y} \) is the final prediction result, and \( \hat{y} \in \{0, 1\} \). Using the frozen CLIP text encoder to extract the feature representation $u$ of $l_{SAP}$, and the CLIP image encoder to extract the feature representation $v$ of the input image.
Next, we compute the contrastive learning loss as follows:
\begin{equation}
\begin{alignedat}{2}
&\mathcal{L}_{v \rightarrow u}
= -\frac{1}{B} \sum_{i=1}^{B}
\log \big( \text{softmax}_j(v_i^{\!T} u_j) \big)_{j=i}, \\
&\mathcal{L}_{u \rightarrow v}
= -\frac{1}{B} \sum_{i=1}^{B}
\log \big( \text{softmax}_j(u_i^{\!T} v_j) \big)_{j=i}, \\
&\mathcal{L}_{\text{contrastive}}
= (\mathcal{L}_{v \rightarrow u} + \mathcal{L}_{u \rightarrow v})/2, \\
\end{alignedat}
\label{eq:contrastive_softmax_split}
\end{equation}

\noindent
In addition, we adopt the cross-entropy loss $\mathcal{L}_{\text{CE}}$ as the classification loss, and the final loss function is obtained by weighting the aforementioned components.
\begin{align}
\mathcal{L}_{\text{loss}} = \mathcal{L}_{\text{contrastive}} + \lambda \mathcal{L}_{\text{CE}}
\end{align}

Notably, across multiple real-world AI-generated image datasets, we were surprised to observe that the scene repository naturally stabilizes, further suggesting that constructing scene-expert models can effectively enhance generalization.

\subsection{Importance-guided Dynamic Optimization}

The SAEM ensures that the detection model maintains strong generalization and adaptability when confronted with diverse scene domains in real-world scenarios. Expert models still suffer from catastrophic forgetting when learning newly emerging generative methods within the same scene. According to gradient projection theory, regardless of the backbone network, if gradient updates occur along directions orthogonal to the subspace spanned by previous features, the model can retain previously learned knowledge during continual learning. The existing detection method with continual learning, RegO \cite{RegO}, which is based on gradient projection theory, achieves competitive performance without relying on a replay buffer. However, due to its inability to perform fine-grained regulation of neurons, it exhibits limited stability and plasticity. Therefore, we introduce the IDOM to mitigate forgetting and maintain stable knowledge across tasks.

First, we employ the Fisher Information Matrix (FIM) to quantify the importance of neurons, as formulated in \cref{eq:fim}

\begin{equation}
\begin{aligned}
\label{eq:fim}
\mathbf{F}_k^{(c)} &= \mathbb{E}\Big[ g_k^{(c)} (g_k^{(c)})^\top \Big|_{\theta=\theta_k^*} \Big]
\end{aligned}
\end{equation}

\noindent
Here, $c \in \{0, 1\}$ indicates whether the sample is \textit{real} (0) or \textit{fake} (1), $g_k^{(c)}$ denotes the gradient of the log-likelihood with respect to parameter $\theta$, and $\theta_k^*$ represents the optimal parameters obtained after training on task $k$. Then, the neuron importance can be calculated as shown in \cref{eq:importance}.

\begin{equation}
\begin{aligned}
\label{eq:importance}
I_k^{(c)}[i][j] = \frac{1}{|S|} \sum_{(s,x)} \frac{1}{N_k^{(s,c)}}{F_k^{(s,c)}[i][j]}
\end{aligned}
\end{equation}

\noindent
where $s\in S$ denotes the scene category of the current input image, $i$, $j$ represents the neuron index and $N_k^{(s,c)}$ represents the number of real or fake samples belonging to scene $s$ in the current task dataset. By setting a threshold $\alpha$, we can identify the core neurons:
\begin{equation}
\begin{aligned}
M_k^{(c)}[i][j] = 
\begin{cases} 
1 & \text{if } I_k^{(c)}[i][j] \geq Q_\alpha \left( I_k^{(c)} \right) \\
0 & \text{otherwise}
\end{cases}
\end{aligned}
\end{equation}

\noindent
Here, \( Q_\alpha(\cdot) \) denotes the $\alpha$-quantile function applied to the importance matrix \( I_k^{(c)} \), which adaptively selects the top $\alpha$ proportion of neurons as core neurons based on their importance scores. \( M_k^{(c)}=1\) denotes neurons that are important for real or fake image detection—defined as core neurons, while \( M_k^{(c)}=0\) indicates neurons that are unimportant for both real and fake image detection—defined as non-core neurons. Through the neuron partitioning mechanism, neurons are divided into core and non-core categories. Starting from the second task, we aggregate all historical task matrices $M$ to obtain a combined matrix $\bar{M}$.

Core neurons are critical for detecting real or fake images. Since fake image features vary significantly across generative methods, the knowledge associated with them is prone to forgetting when learning new tasks. In contrast, real image features are more compact and stable. Therefore, when updating core neurons, it is crucial to strictly preserve the knowledge related to real-image detection acquired from previous tasks while minimizing the forgetting of fake-image detection capabilities. This ensures that the model can learn new generative patterns without compromising its existing detection ability. To achieve this, an importance-guided gradient projection strategy is applied.

For neurons important to real image detection, the current gradients $g$ are projected onto the directions of the old task gradients $\hat{g}$, producing the parallel component $g_p$ that preserves previously learned knowledge. In contrast, for neurons important to fake image detection, the gradients are adjusted to be orthogonal to those from previous tasks, yielding the orthogonal component $g_o$ that enables the model to learn new generative patterns without interfering with prior representations.
\begin{equation}
\begin{aligned}
g_{p} &= \frac{g^\top \hat{g}}{\|\hat{g}\|^2} \cdot \hat{g},&g_{o} = g-g_{p}
\end{aligned}
\end{equation}

The historical importance of each neuron for real and fake images is further aggregated, denoted as $\tilde{I}_k^{(c)} = \sum_{t=1}^{k-1} I_t^{(c)}$. Their relative ratio is then used to compute a control factor that adaptively determines the neuron’s update direction, as shown in \cref{eq:qfactor}.

\begin{equation}
\begin{aligned}
\label{eq:qfactor}
q_{0} &= \frac{\tilde{I}_{k}^{(0)}}{\tilde{I}_k^{(0)} + \tilde{I}_k^{(1)}}, \quad
q_{1} = 1 - q_{0}.
\end{aligned}
\end{equation}

To further prevent drastic updates to highly important neurons, we introduce an importance-based gradient scaling function:
\begin{equation}
\begin{aligned}
u_k = \frac{1}{1 + e \cdot \bar{I}_i},
\end{aligned}
\end{equation}

\noindent
where $\bar{I}_i$ denotes the normalized historical importance and $e$ controls the suppression strength. Finally, the overall optimization for core neurons is formulated as:
\begin{equation}
\begin{aligned}
\label{eq:core}
\widetilde{g} &= q_{0} \cdot g_{p} + q_{1} \cdot g_{o}, \\
g_A &= u_k \cdot \widetilde{g} \odot \mathbb{I}_{{\overline{\mathbf{M}}[i][j] = 1}}.
\end{aligned}
\end{equation}

\noindent
where $\mathbb{I}_{{\overline{\mathbf{M}}[i][j] = 1}}$ is an indicator function that takes the value 1 when $\overline{M}$ = 1, and 0 otherwise.
After handling core neurons, we address non-core neurons to ensure that the model maintains sufficient plasticity during continual learning. For non-core neurons, we allow them to be freely updated as:
\begin{equation}
\begin{aligned}
g_B = g \odot \mathbb{I}_{{\overline{\mathbf{M}}[i][j] = 0}},
\end{aligned}
\end{equation}

Thus, the overall gradient update for all neurons is defined as:
\begin{equation}
\begin{aligned}
w = g_A + g_B.
\end{aligned}
\end{equation}

\section{Experiment}

In this section, we conduct a comprehensive evaluation to assess the practical effectiveness of different methods, including continual learning performance, generalization across open-world datasets, robustness under common image degradation, and ablation studies.


\subsection{Experimental Settings}
\begin{table*}[ht]
    \centering
    \footnotesize
    \caption{Performance evaluation and comparison with other methods under Protocol-1 (\%), with the best results highlighted in bold and the second-best results underlined.}
    \label{tab:performance_comparison_in_continual_learning}

\begin{subtable}{\textwidth}
\centering

\renewcommand{\arraystretch}{0.9}  
\resizebox{\textwidth}{!}{
\begin{tabular}{c c c c *{6}{cc}}
    \toprule
    \multirow{2}{*}{Method} 
    & \multirow{2}{*}{Venue}  
    & \multirow{2}{*}{\shortstack{Continual\\Learning}} 
    & \multirow{2}{*}{\shortstack{Replay\\Set}}
    & \multicolumn{1}{c}{1-ADM} 
    & \multicolumn{2}{c}{2-GLIDE} 
    & \multicolumn{2}{c}{3-SAGAN}
    & \multicolumn{2}{c}{4-ProGAN} \\
    \cmidrule(lr){5-11}
    & & & & AA & AA & AF & AA & AF & AA & AF \\ 
    \midrule
    CLIP+LoRA & - & $\times$ & $-$ & \textbf{99.95} & 98.42 & 3.01 & 90.83 & 13.68 & \underline{98.44} & \underline{1.60}  \\
    \midrule
    Universe \cite{2023clipdetection} & CVPR'23 & $\times$ & $-$ & 96.10 & 89.41 & 10.66 & 87.01 & 9.76 & 88.48 & 6.73  \\
    NPR \cite{2024NPR} & CVPR'24 & $\times$ & $-$ & 93.65 & 92.36 & \textbf{-0.55} & 79.05 & 21.91 & 82.89 & 14.46  \\
    \midrule
    RegO \cite{RegO} & AAAI'25 & $\checkmark$ & $\times$ & 99.58 & 98.09 & 0.88 & 94.09 & 7.22 & 97.14 & 1.66 \\
    Tang et al. \cite{tifs} & TIFS'25 & $\checkmark$ & $\checkmark$ & 99.21 & \underline{99.10} & 0.18 & \underline{97.50} & \underline{2.12} & 94.58 & 4.32 \\
    \rowcolor{cyan!10}  
    SAIDO (Ours) & - & $\checkmark$ & $\times$ & \underline{99.81} & \textbf{99.54} & \underline{0.06} & \textbf{99.41} & \textbf{0.42} & \textbf{98.67} & \textbf{0.56} \\
    \bottomrule
\end{tabular}}
\renewcommand{\arraystretch}{1} 
\end{subtable}

\vspace{0.4em}

\begin{subtable}{\textwidth}
\centering
\resizebox{\textwidth}{!}{
\begin{tabular}{c *{10}{cc} c}
    \toprule
    \multirow{2}{*}{Method} 
    & \multicolumn{2}{c}{5-BigGAN}
    & \multicolumn{2}{c}{6-Wukong} 
    & \multicolumn{2}{c}{7-SD1.5}
    & \multicolumn{2}{c}{8-VQDM} 
    & \multicolumn{2}{c}{9-Midjourney-V5}
    & \multirow{2}{*}{\shortstack{New.\\ACC~($\uparrow$)}} \\ 
    \cmidrule(lr){2-11}
    & AA & AF & AA & AF & AA & AF & AA & AF & AA & AF & \\ 
    \midrule
    CLIP+LoRA & 96.06 & 4.31 & 92.96 & 7.89 & 91.60 & 9.32 & 87.80 & 13.52 & 89.63 & 11.17 & \textbf{99.56} \\
    \midrule
    Universe \cite{2023clipdetection} & 85.42 & 9.73 & 85.60 & 9.63 & 84.55 & 10.39 & 83.92 & 11.22 & 80.59 & 14.46 & 93.45 \\
    NPR \cite{2024NPR} & 70.44 & 29.07 & 75.18 & 22.76 & 74.05 & 24.17 & 67.94 & 31.17 & 81.92 & 15.17 & 95.40 \\
    \midrule
    RegO \cite{RegO} & \underline{96.63} & \underline{2.40} & 92.31 & 6.59 & 90.96 & 7.52 & 92.55 & 5.24 & 86.34 & 11.30 & 90.04 \\
    Tang et al. \cite{tifs} & 94.20 & 4.95 & \underline{94.04} & \underline{4.46} & \underline{91.87} & \underline{7.06} & \underline{93.46} & \underline{5.10} & \underline{92.13} & \underline{6.63} & 92.88 \\
    \rowcolor{cyan!10}  
    SAIDO (Ours) & \textbf{98.50} & \textbf{0.95} & \textbf{97.92} & \textbf{1.52} & \textbf{97.45} & \textbf{2.02} & \textbf{98.10} & \textbf{1.29} & \textbf{95.61} & \textbf{3.94} & \underline{97.27} \\
    \bottomrule
\end{tabular}}
\end{subtable}
\end{table*}

\begin{figure*}[t]
    \centering
    \includegraphics[width=0.85\linewidth]{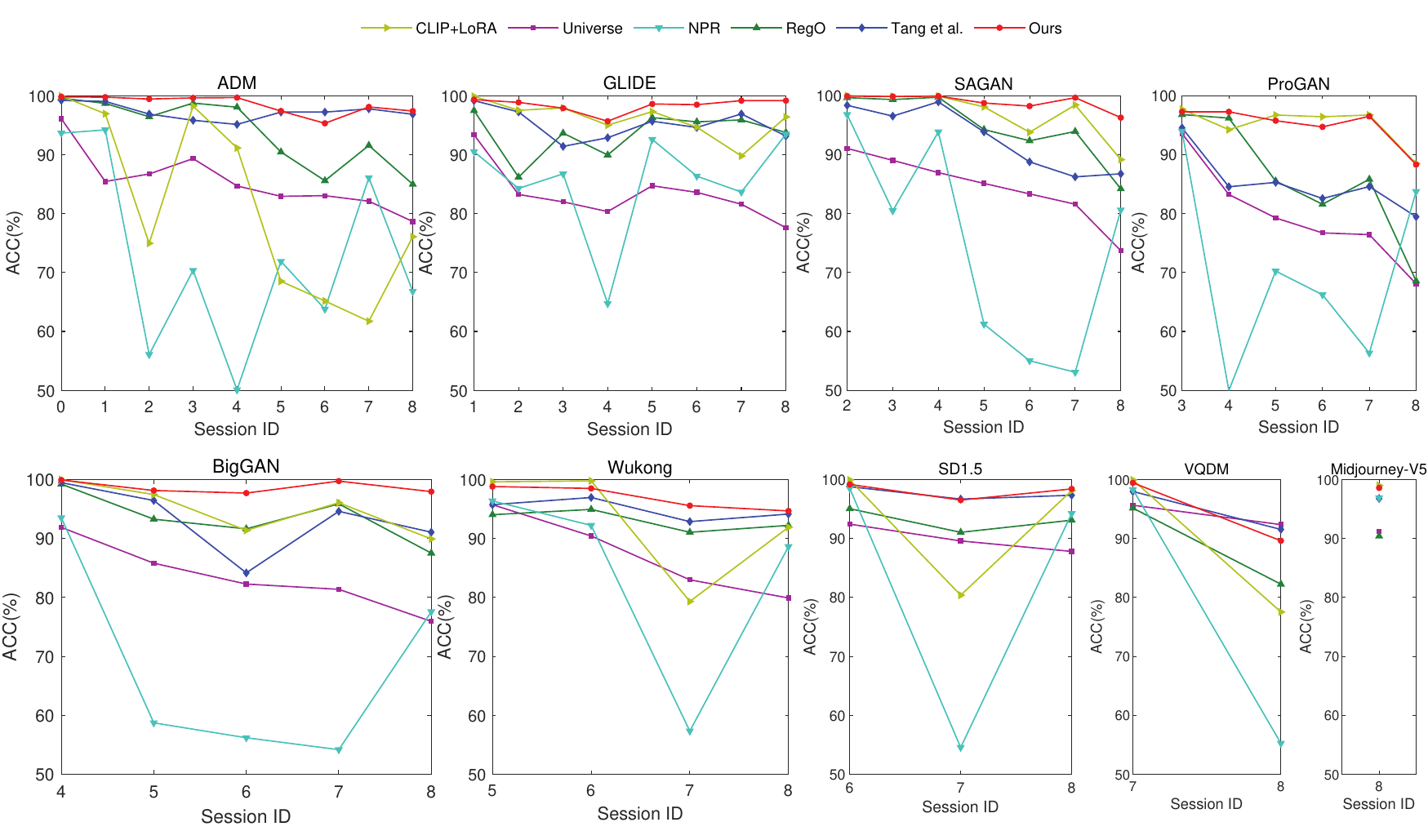} 
    \caption{Detailed performance of each method on Protocol 1. The line charts show the performance on each generative dataset, where the lines show the performance
variation of different methods on a particular dataset across different sessions.}
    \label{fig:performance evaluation}
\end{figure*}

\begin{table*}[t]
    \centering
    \caption{Generalization performance comparison on open-world dataset under Protocol 2 (\%).}
    \label{tab:generalization_comparison}
    \begin{tabular}{c *{7}{c}}
        \toprule
        \multirow{2}{*}{\centering Method} 
        & \multicolumn{6}{c}{Out-of-Distribution (OOD) Tasks} 
        & \multirow{2}{*}{\centering{AA}} \\ 
        \cmidrule(lr){2-7}
        & StyleGAN-xl & R3GAN & FLUX1-dev & Midjourney-V6 & SD3 & Imagen3 & \\ 
        \midrule
        Universe \cite{2023clipdetection}
        & 60.22 & 50.72 & 73.27 & 65.92 & 65.53 & 61.48 & 77.57 \\
        NPR \cite{2024NPR} \ & 90.32 & 50.53 & \underline{87.41} & \underline{90.32} & \underline{84.92} & \underline{81.38} & 80.81 \\
        \midrule
        RegO \cite{RegO}      
        & 92.00 & 81.75 & 72.52 & 88.54 & 72.17 & 70.89 & 79.65 \\
        Tang et al. \cite{tifs}  
        & \textbf{98.00} & \underline{87.01} & 74.50 & 88.15 & 77.50 & 68.21 & \underline{81.88} \\
        \rowcolor{cyan!10}  
        SAIDO (Ours)    
        & \underline{96.94} & \textbf{97.83} & \textbf{88.10} & \textbf{94.89} & \textbf{87.85} & \textbf{82.50} & \textbf{91.35} \\
        \bottomrule
    \end{tabular}
    
\end{table*}

\textbf{Datasets.} In experiments, we employed a diverse collection of AI-generated image datasets encompassing both classical and advanced generative models. Specifically, nine image generators were considered in the continual learning tasks, including ProGAN \cite{karras2018progressive}, SAGAN \cite{SAGAN}, BigGAN \cite{brock2018large}, ADM \cite{ADM}, GLIDE \cite{glide}, Wukong \cite{wukong2022}, SD1.5 \cite{SD1.5}, VQDM \cite{VQDM}, and Midjourney-V5 \cite{midjourney2022}. Images generated by ProGAN and BigGAN are obtained from CNNSpot \cite{cnnspot}, while those produced by SAGAN are sourced from OSMA \cite{OSMA}. The images generated by all remaining models are collected from GenImage \cite{genimage}. For open-world generalization evaluation, six advanced generators are involved, including StyleGAN-xl \cite{stylegan-xl}, R3GAN \cite{R3GAN}, FLUX1-dev \cite{flux}, Midjourney-V6 \cite{midjourneyv6}, SD3 \cite{SD3}, and Imagen3 \cite{imagen3}. Images from StyleGAN-xl and R3GAN are generated by the authors’ pretrained models on ImageNet, while others were obtained from AIGIBench \cite{AIGIBench}. 

\noindent
\textbf{Metrics.} The evaluation metrics used in this study include Accuracy (ACC), Average Accuracy (AA), Average Forgetting (AF), and New Accuracy (New.ACC). ACC indicates the proportion of correctly predicted samples in each task, while AA and AF are used to evaluate the overall detection performance and memory stability throughout the continual learning process, respectively. New.ACC denotes the accuracy evaluated on the current training dataset at each incremental step, and the final score is obtained by averaging these values, reflecting the method’s plasticity.

\noindent
\textbf{Protocols.} To systematically analyze the performance of different methods in continual learning tasks and their generalization ability when facing unknown generative methods in open-world scenarios, two experimental protocols were established. To ensure the rigor of our experiments, we also set multiple dataset orders, and the corresponding results are presented in the supplementary material.

\begin{itemize}
    \item \textbf{Protocol 1.} This protocol constructs a sequential dataset $\mathcal{D}$ = \{ADM, GLIDE, SAGAN, ProGAN, BigGAN, Wukong, SD1.5, VQDM, Midjourney-V5\} to evaluate the performance of different methods in continual learning tasks.
    \item \textbf{Protocol 2.} This protocol builds an open-world dataset $\mathcal{T}$ = \{StyleGAN-xl, R3GAN, FLUX1-dev, Midjourney-V6, SD3, Imagen3\}  to evaluate the generalization ability of different methods in an open-world setting.
\end{itemize}

\begin{table*}[t]
  \centering
  \caption{Performance of Different Detectors under Robust Settings (\%). AA1 denotes the average detection accuracy on generative models from Protocol 1, while AA2 denotes the average detection accuracy on generative models from Protocol 2.}
  \label{table:detector_robust}
  \begin{tabular}{lcc cc cc cc cc}  
    \toprule
    \multirow{2}{*}{\begin{tabular}[c]{@{}c@{}}Detectors $\rightarrow$\\ Robust Settings $\downarrow$\end{tabular}} 
    & \multicolumn{2}{c}{Universe \cite{2023clipdetection}} 
    & \multicolumn{2}{c}{NPR \cite{2024NPR}} 
    & \multicolumn{2}{c}{RegO \cite{RegO}}  
    & \multicolumn{2}{c}{Tang et al. \cite{tifs}} 
    & \multicolumn{2}{c}{SAIDO (Ours)} \\ 
    \cmidrule(lr){2-3} \cmidrule(lr){4-5} \cmidrule(lr){6-7} \cmidrule(lr){8-9} \cmidrule(lr){10-11} 
    & AA1 & AA2 
    & AA1 & AA2 
    & AA1 & AA2 
    & AA1 & AA2 
    & AA1 & AA2  \\
    \midrule
    Origin 
    & 55.36 & 62.68 
    & 64.62 & 61.87 
    & 83.63 & \underline{80.62} 
    & \underline{91.35} & 75.39 
    & \textbf{93.24} & \textbf{81.62} \\
    JPEG Compression 
    & 51.80 & 53.92
    & 63.70 & 50.08 
    & \underline{87.43} & \textbf{78.20}
    & 77.52 & 68.83 
    & \textbf{92.27} & \underline{77.81} \\
    Gaussian Noise 
    & 51.15 & 51.56
    & 57.30 & 50.36  
    & \underline{82.25} & \underline{83.71} 
    & \textbf{85.31} & 69.49 
    & 79.61 & \textbf{84.12} \\
    Up-down Sampling 
    & 52.30 & 54.59
    & 71.73 & 50.51  
    & \underline{85.51} & \underline{81.76}
    & 84.29 & 76.64 
    & \textbf{90.03} & \textbf{82.03} \\
    \midrule
    Mean 
    & 52.65 & 55.69 
    & 64.25 & 53.20
    & \underline{84.71} & \underline{81.07}
    & 84.62 & 72.59 
    & \textbf{88.79} & \textbf{81.40} \\
    \bottomrule
  \end{tabular}
\end{table*}

\begin{table*}[t]
    \centering
    \footnotesize
    \caption{Ablation study of different components (\%). RAO represents the gradient projection update strategy of RegO.}
    \label{tab:performance_comparison_ablation}
    \resizebox{\textwidth}{!}{
    \begin{tabular}{c c *{8}{cc}}  
        \toprule
        \multirow{2}{*}{Method}  
        & \multicolumn{1}{c}{ADM} 
        & \multicolumn{2}{c}{GLIDE} 
        & \multicolumn{2}{c}{SAGAN}
        & \multicolumn{2}{c}{ProGAN} 
        & \multicolumn{2}{c}{BigGAN}
        & \multicolumn{2}{c}{Wukong} 
        & \multicolumn{2}{c}{SD1.5}
        & \multicolumn{2}{c}{VQDM} 
        & \multicolumn{2}{c}{Midjourney-V5} \\
        \cmidrule(lr){2-18}  
        & AA & AA & AF & AA & AF & AA & AF & AA & AF & AA & AF & AA & AF & AA & AF & AA & AF \\  
        \midrule
        CLIP+SAEM   
        & \underline{99.81}  
        & \underline{99.39} & \underline{0.65} & 96.21 & 5.36 & 95.06 & 5.65 & 97.07 & 2.85 & 93.20 & 7.27 & 93.39 & 6.72 & 95.11 & 4.54 & 88.85 & 11.41 \\
        CLIP+IDOM   
        & \textbf{99.95} 
        & 99.24 & 1.30 & \underline{98.55} & \underline{2.08} & \textbf{99.01} & \textbf{0.42} & \textbf{98.50} & \underline{1.20} & \underline{97.45} & \underline{2.43} & \underline{97.31} & \underline{2.54} & \textbf{98.87} & \textbf{0.73} & \underline{94.81} & \underline{4.60} \\
        CLIP+SAEM+RAO   
        & \underline{99.81} 
        & 99.12 & 0.51 & 96.84 & 4.11 & 96.33 & 3.62 & 97.79 & 1.79 & 96.61 & 3.00 & 96.88 & 2.57 & 97.07 & 2.30 & 92.89 & 6.88 \\
        \rowcolor{cyan!10}  
        SAIDO (Ours)   
        & \underline{99.81} 
        & \textbf{99.54} & \textbf{0.06} & \textbf{99.41} & \textbf{0.42} & \underline{98.67} & \underline{0.56} & \textbf{98.50} & \textbf{0.95} & \textbf{97.92} & \textbf{1.52} & \textbf{97.45} & \textbf{2.02} & \underline{98.10} & \underline{1.29} & \textbf{95.61} & \textbf{3.94} \\
        \bottomrule
    \end{tabular}%
    }
\end{table*}

\noindent
\textbf{Methods.}
The proposed method is evaluated against SOTA detection methods with continual learning, including RegO \cite{RegO} and Tang et al. \cite{tifs}, and further compared with methods without continual learning through multi-level performance analyses, including Universe \cite{2023clipdetection} and NPR \cite{2024NPR}. In addition, LoRA fine-tuning on CLIP is used as an approximate upper bound for assessing model plasticity. All methods are evaluated strictly using their official implementations under the same experimental settings. Each experiment is conducted multiple times, and the final results are reported as the average values.

\noindent
\textbf{Implementation Details.}
In our framework, the ViT-L/14 variant of CLIP is adopted as the backbone. We employ the SGD optimizer with momentum, using an initial learning rate of 0.01 and training for 10 epochs. In our experiments, the hyperparameter $\alpha$ is set to 0.75, and $e$ is set to 1.0.

\subsection{Performance Evaluation on Continual Learning Tasks}
Continual learning performance is evaluated on the sequential dataset $\mathcal{D}$ defined in Protocol 1. To ensure a fair comparison, we distinguish whether each method utilizes a replay set. As shown in \cref{tab:performance_comparison_in_continual_learning}, even without the support of a replay buffer, our method achieves comprehensive improvements in both AA and AF compared to current SOTA approaches. Specifically, in terms of stability, the proposed method achieves an AA of 95.61\% and an AF of 3.94\%. Compared to the second-best approach, it yields relative reductions of 44.22\% in error rate and 40.57\% in average forgetting. Regarding plasticity, the method attains a New Accuracy of 97.27\%, demonstrating its superior ability to balance stability and plasticity. The performance improvement mainly arises from two aspects. On one hand, the IDOM performs fine-grained dynamic optimization of neurons, effectively balancing model plasticity and stability without relying on data replay. On the other hand, the SAEM adopts an extensible independent scene-expert architecture, enabling the detection model to maintain stability as new scene domains continuously emerge in continual learning tasks.

\subsection{Performance Evaluation on Open-World Tasks}
In this subsection, we simulate an open-world scenario following Protocol 2, where images generated by six unseen advanced generative models are used to construct open-world testing tasks. The SAIDO and other approaches are evaluated in terms of their generalization ability under this open-world setting. To ensure fair training, all models are trained on the dataset specified in Protocol 1. As shown in \cref{tab:generalization_comparison}, our method surpasses existing SOTA approaches in average accuracy on open-world tasks, outperforming the second-best method by 9.47\%. This improvement can be attributed to both the SAEM and IDOM. The SAEM module, with its scene experts, captures forgery-relevant cues within particular scene domains, enhancing generalization to unseen generative methods in the open world. Meanwhile, the IDOM module implements fine-grained control over neuron updates, enabling the model to continuously improve generalization throughout continual learning.

\subsection{Robustness Evaluation}
In real-world applications, there is a strong demand for detection methods that can effectively adapt to various types of image degradations. We evaluate the robustness of different methods under three common degradation types: JPEG compression, Gaussian noise, and Up-down sampling. The degradation settings follow the robustness evaluation protocol in AIGIBench \cite{AIGIBench}. Specifically, the JPEG compression quality factor is set to 50, the standard deviation of the pixel-wise Gaussian noise to 4, and for up-down sampling, the image is reduced to half its original size using the nearest neighbor algorithm and then up-sampled back to the original resolution.

To enhance robustness, data augmentation is applied during training for all methods, where JPEG compression and Gaussian noise are introduced with a probability of 0.5, following the default configuration in CNNSpot \cite{cnnspot}. As shown in \cref{table:detector_robust}, SAIDO achieves the best performance in robustness tests against both known and unseen generative models. This demonstrates that our approach can better learn features augmented by data degradations, effectively balancing plasticity and stability in real-world scenarios.

\subsection{Ablation Study}

The effectiveness of each component, namely SAEM and IDOM, is evaluated. Notably, IDOM serves as a key module in our framework, enhancing model plasticity and memory stability in continual learning tasks. To further validate the adaptability and superiority of IDOM within our framework, we conduct additional comparison experiments with RAO, the neuron update strategy based on gradient projection from the method RegO \cite{RegO}. All experiments are conducted under Protocol 1.

As shown in \cref{tab:performance_comparison_ablation}, under identical experimental settings, IDOM consistently outperforms RAO in both average detection accuracy and average forgetting rate across all tasks, except for the first one where their performances are identical. Moreover, when both SAEM and IDOM are jointly employed, our framework achieves the best overall performance. It is worth noting that when SAEM is disabled and only IDOM is used, the model still attains optimal results on certain tasks. We attribute this to the fact that when the scene-domain distributions among different tasks are relatively concentrated, the scene-domain shifts are mild, limiting the advantages of SAEM, while training more data within a single LoRA model can slightly boost performance.

\section{Conclusion}
In this paper, a novel continual learning framework for AI-generated image detection, SAIDO, is proposed. SAIDO consists of two key components: SAEM and IDOM. SAEM leverages VLLMs for scene awareness to dynamically manage LoRA scene expert modules, enhancing generalization across diverse real-world scenes. IDOM employs importance-guided gradient projection to fine-tune neuron update directions, effectively balancing model plasticity and stability. Extensive experiments show that SAIDO outperforms current SOTA methods, achieving relative reductions of 44.22\% and 40.57\% in average error rate and forgetting rate, respectively, and improving average detection accuracy on open-world datasets by 9.47\%.
{
    \small
    \bibliographystyle{ieeenat_fullname}
    \bibliography{main}
}

\clearpage
\setcounter{page}{1}
\maketitlesupplementary

\section{Theoretical Proof}

This section provides a theoretical analysis of the gradient-projection strategy adopted in IDOM, demonstrating its effectiveness in continual learning. Assume that the training of task $t$ has been completed, and task $t+1$ is about to be trained. Let the inputs from the previous task be denoted as $x_t$, and let the trainable neurons for the new and old tasks be represented as $E_{t+1}$ and $E_{t}$, respectively. First, consider the case where a neuron is solely relevant to fake image detection. During the training of the new task, the goal is to prevent forgetting of the old task, which requires that the condition in \cref{eq:forgetting} should hold.

\begin{equation}
f_{\phi + \phi_n}(E_{t+1}, x_t) = f_{\phi + \phi_n}(E_t, x_t),
\label{eq:forgetting}
\end{equation}

\noindent
For $E_{t+1}$, it is obtained by updating $E_t$ after training on task $t$. Let $\Delta E_1$ denote the gradient update of the neurons after training on task $t$; then, \cref{eq:17} can be derived.

\begin{equation}
\label{eq:17}
E_{t+1} = E_t + \Delta E_1
\end{equation}

\noindent
Then, if the following equation holds, forgetting can be avoided.

\begin{equation}
\label{eq:avoided}
E_{t+1}x_t = E_tx_t
\end{equation}

\noindent
By substituting \cref{eq:17} into \cref{eq:avoided}, the following equation can be further obtained.

\begin{equation}
\label{eq:final forgetting}
x_t\Delta E_1 = 0
\end{equation}

\noindent
\cref{eq:final forgetting} indicates that if neurons are updated along directions orthogonal to the subspace spanned by the features of previous tasks, forgetting is significantly reduced.

Assuming that the neurons are only related to real image detection, according to our previous analysis, real images exhibit a more compact feature distribution. Therefore, unlike fake images, real images from different tasks are treated as a single class during detection, denoted as $x_{t+1} \approx x_t$. Therefore, during neuron updates, greater emphasis is placed on preserving knowledge from previous tasks by updating neurons along the direction of the projection of the current gradient onto the feature subspace of previous tasks.

\begin{equation}
\label{eq:real}
E_{t+1} = E_t + \Delta E_0^t
\end{equation}

\noindent
According to \cref{eq:real}, after completing the $t$-th training task, $\mathcal{L}{\text{loss}}(\Delta E_0^t, x_t)$ reaches its minimum. When the neuron is updated along the projection direction of the current gradient onto the feature subspace of the old tasks, and considering $x_{t+1} \approx x_t$ under the gradient projection strategy, it follows that $\Delta E_0^{t+1}=\beta E_0^t$, where $\beta$ denotes a scaling coefficient. Consequently, $\mathcal{L}_{\text{loss}}(\Delta E_0^{t+1}, x{t+1})$ also attains its minimum.

We introduce the Fisher matrix to measure neuron importance, characterizing each neuron's contribution to real-image or fake-image detection. Empirically, it is observed that during network updates, almost no neuron is entirely irrelevant to either real or fake image detection. The current SOTA method RegO \cite{RegO} attempts to divide neurons using multiple importance thresholds, but such discrete partitioning fails to provide precise control over neuron updates, limiting both plasticity and stability in continual learning. Therefore, aggregated neuron importance is used to determine the proportional relationship between update directions, enabling finer-grained regulation of how each neuron is updated.

\begin{equation}
\begin{aligned}
\label{eq:imp_x}
q_{0} = &\frac{\tilde{I}_{k}^{(0)}}{\tilde{I}_k^{(0)} + \tilde{I}_k^{(1)}}, \quad
q_{1} = 1 - q_{0}.
\end{aligned}
\end{equation}

Based on the theoretical analysis, $q_0$ and $q_1$ are used to assign different proportions of the update to the projection of the gradient $g_{p}$ onto the old-task feature subspace and its orthogonal component $g_{o}$. This enables precise control over neuron update directions according to their importance.

\begin{equation}
\begin{aligned}
\label{eq:core_x}
\widetilde{g} &= q_{0} \cdot g_{p} + q_{1} \cdot g_{o}
\end{aligned}
\end{equation}

\section{Dataset Details}
To ensure the rigor of the experiments, this section provides a detailed discussion of the dataset setups in both continual learning tasks and open-world tasks, along with the scene distributions in each dataset.
\subsection{Continual Learning Dataset}

During the training of continual learning tasks, taking the task-order configuration of Protocol 1 as an example, SAEM first employs multiple rounds of large-model scoring to select the highest-scoring set of base scenes, forming the initial scene configuration $S_{initial}$ = \{Activity, Animal, Building, Food, Nature, Object, Person, Vehicle\}. Throughout training, the VLLM identifies scene features in the ADM dataset and updates the scene set with $Plant$ and $Clothing$. As subsequent tasks arrive, the scene updates gradually stabilize, resulting in the final scene library: $S_{final}$ = \{Activity, Animal, Building, Clothing, Food, Nature, Object, Person, Plant, Vehicle\}

\begin{figure}[H] 
    \centering 
    \includegraphics[width=\linewidth]{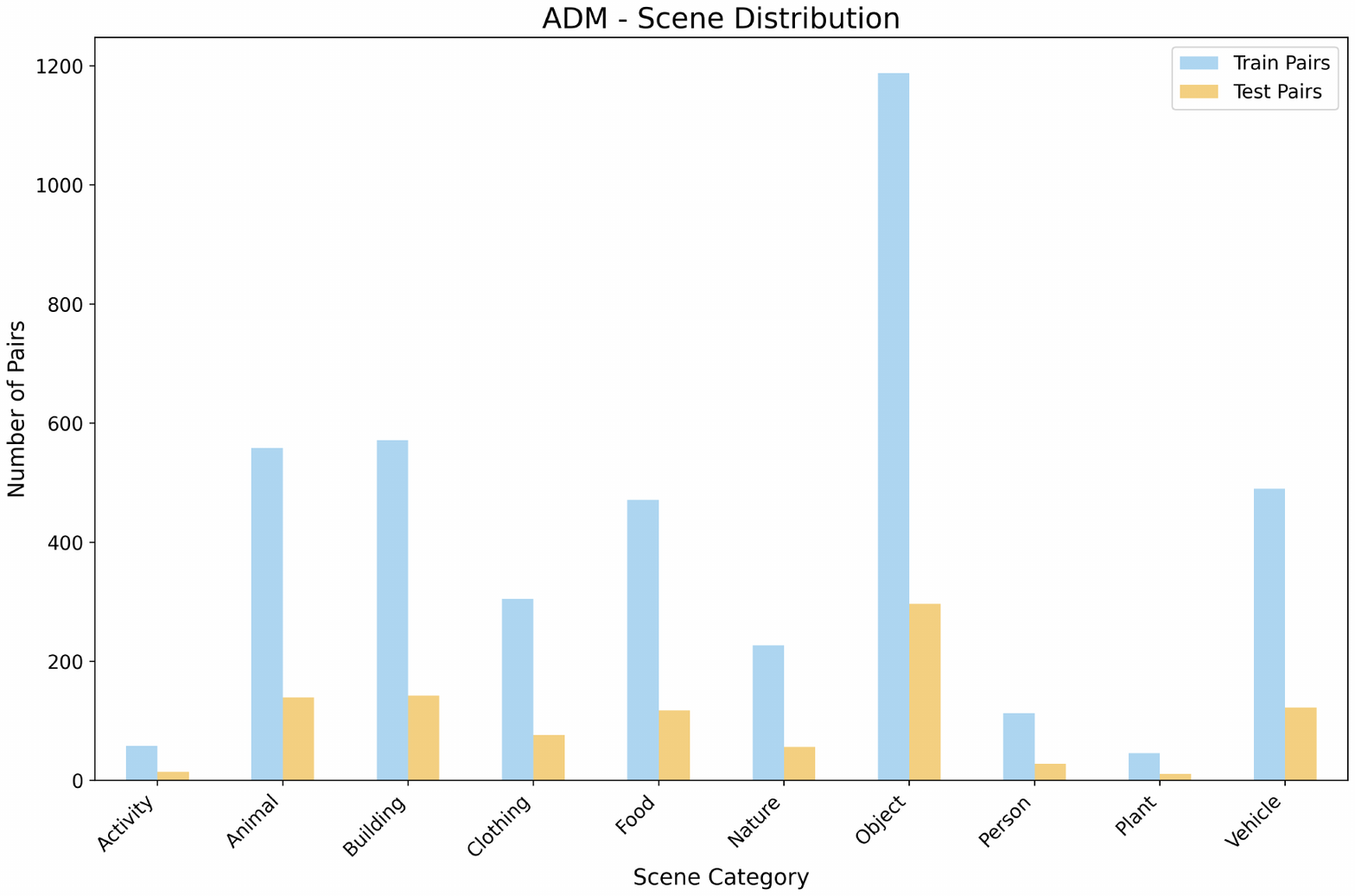} 
    \caption{The scene distribution of ADM}
    \label{ADM} 
\end{figure}

\textbf{ADM \cite{ADM}.} We randomly select 4K sample pairs from the ADM dataset provided by GenImage \cite{genimage} as the training set. The test set size for the entire continual learning task is consistently 25\% of the training set size. To ensure the diversity of image content, real images are selected from the large-scale ImageNet dataset, while the generative models are trained using ImageNet. The specific scene distribution in the dataset is shown in \cref{ADM}.

\begin{figure}[H] 
    \centering 
    \includegraphics[width=\linewidth]{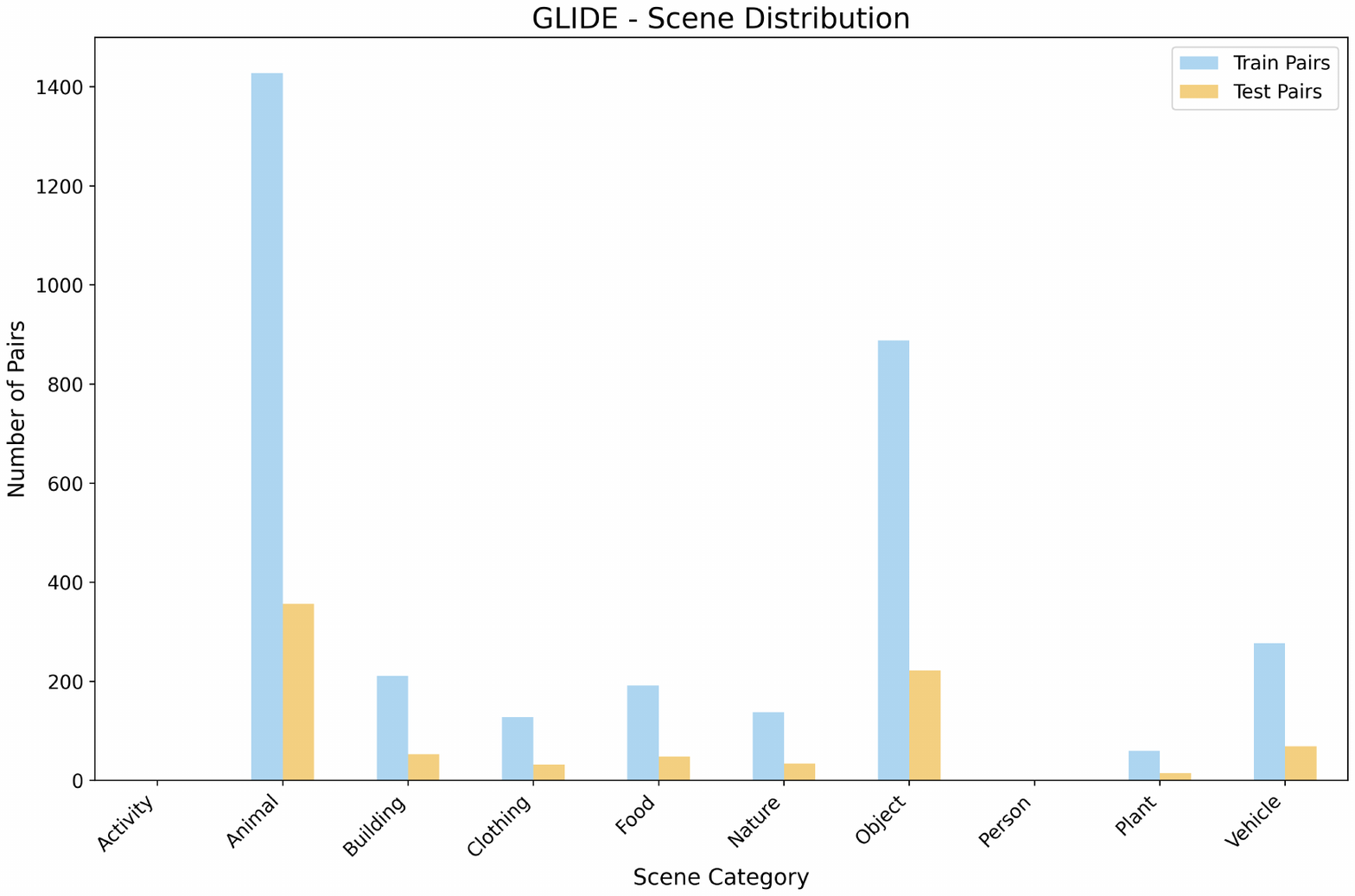} 
    \caption{The scene distribution of GLIDE}
    \label{GLIDE} 
\end{figure}

\noindent
\textbf{GLIDE \cite{glide}.} 3K sample pairs from the GLIDE subset of GenImage \cite{genimage} are used as the training data. The generative model is trained on ImageNet. The scene distribution is shown in \cref{GLIDE}.

\begin{figure}[H] 
    \centering 
    \includegraphics[width=\linewidth]{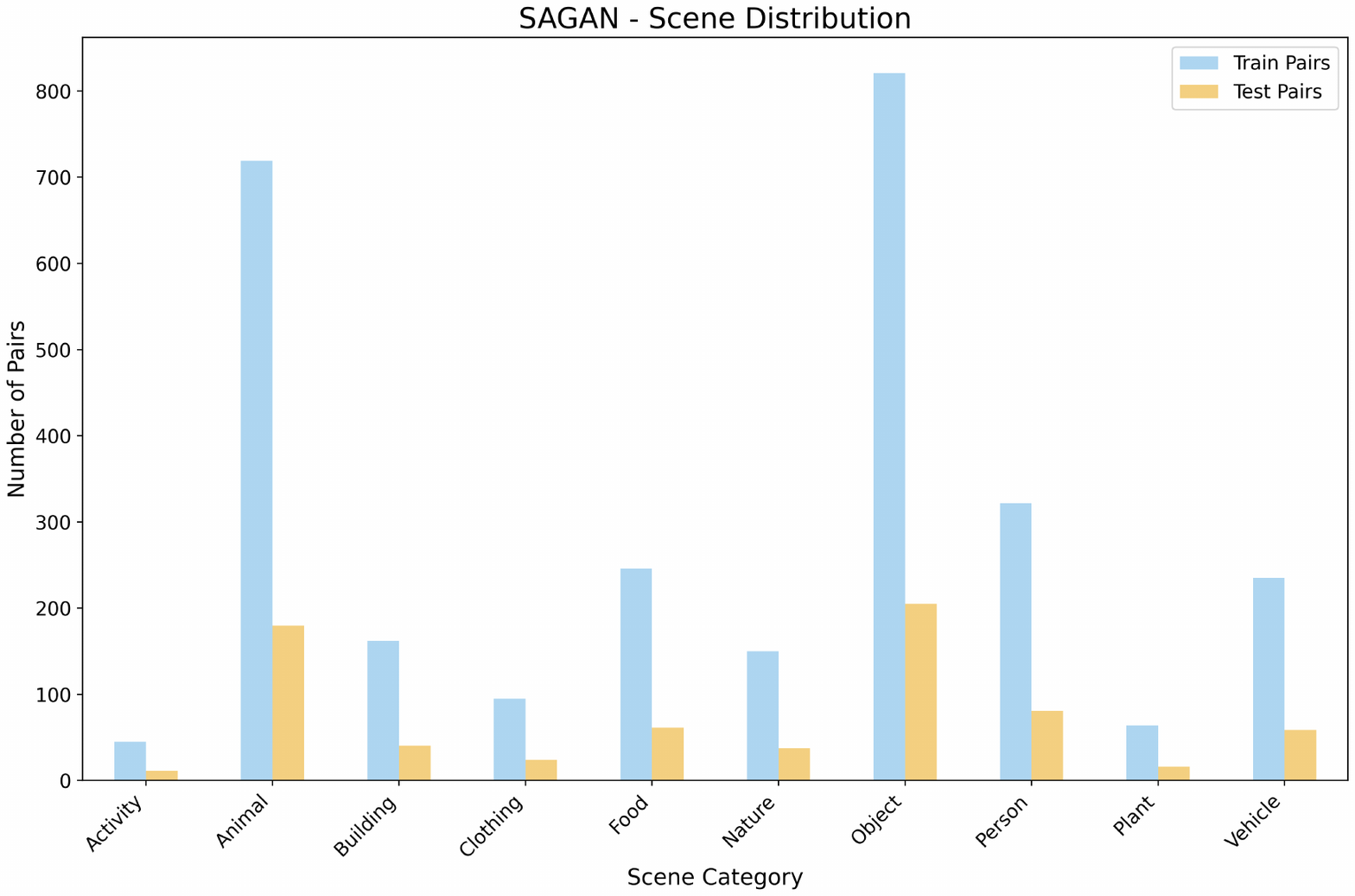} 
    \caption{The scene distribution of SAGAN}
    \label{SAGAN} 
\end{figure}

\noindent
\textbf{SAGAN \cite{SAGAN}.} 3K training pairs are drawn from the SAGAN dataset in OSMA \cite{OSMA}, using ImageNet for both real images and generative model training. The scene distribution is in \cref{SAGAN}.

\begin{figure}[H] 
    \centering 
    \includegraphics[width=\linewidth]{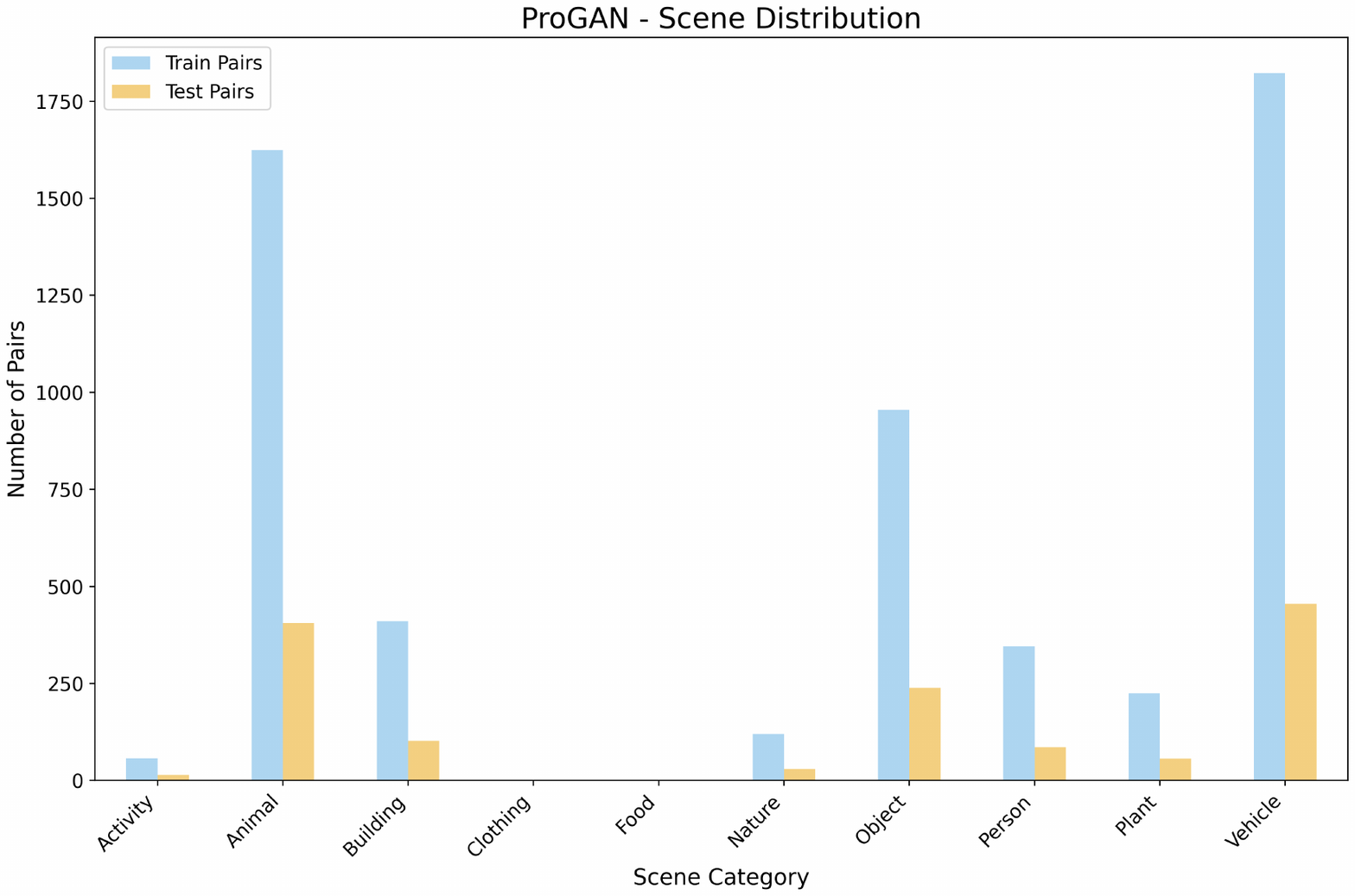} 
    \caption{The scene distribution of ProGAN}
    \label{ProGAN} 
\end{figure}

\noindent
\textbf{ProGAN \cite{karras2018progressive}.} 6K pairs from the ProGAN dataset in CNNSpot \cite{cnnspot} are used. LSUN provides the real images and serves as the training source for the generative model. The scene distribution is shown in \cref{ProGAN}.

\begin{figure}[H] 
    \centering 
    \includegraphics[width=\linewidth]{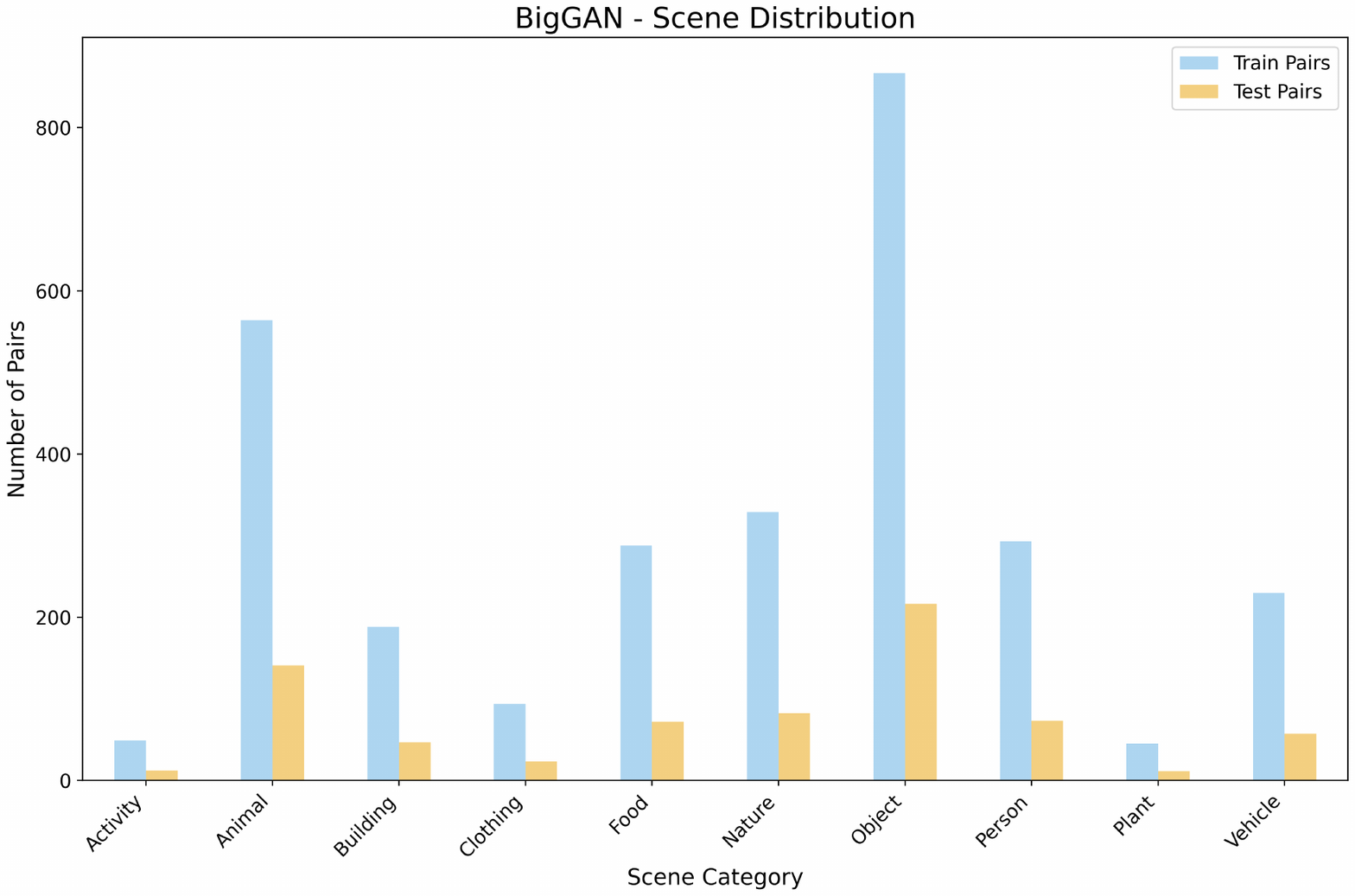} 
    \caption{The scene distribution of BigGAN}
    \label{BigGAN} 
\end{figure}

\noindent
\textbf{BigGAN \cite{brock2018large}.} 3K sample pairs from CNNSpot \cite{cnnspot} form the training set, with ImageNet used for both real images and generative model training. The scene distribution is in \cref{BigGAN}.

\begin{figure}[H] 
    \centering 
    \includegraphics[width=\linewidth]{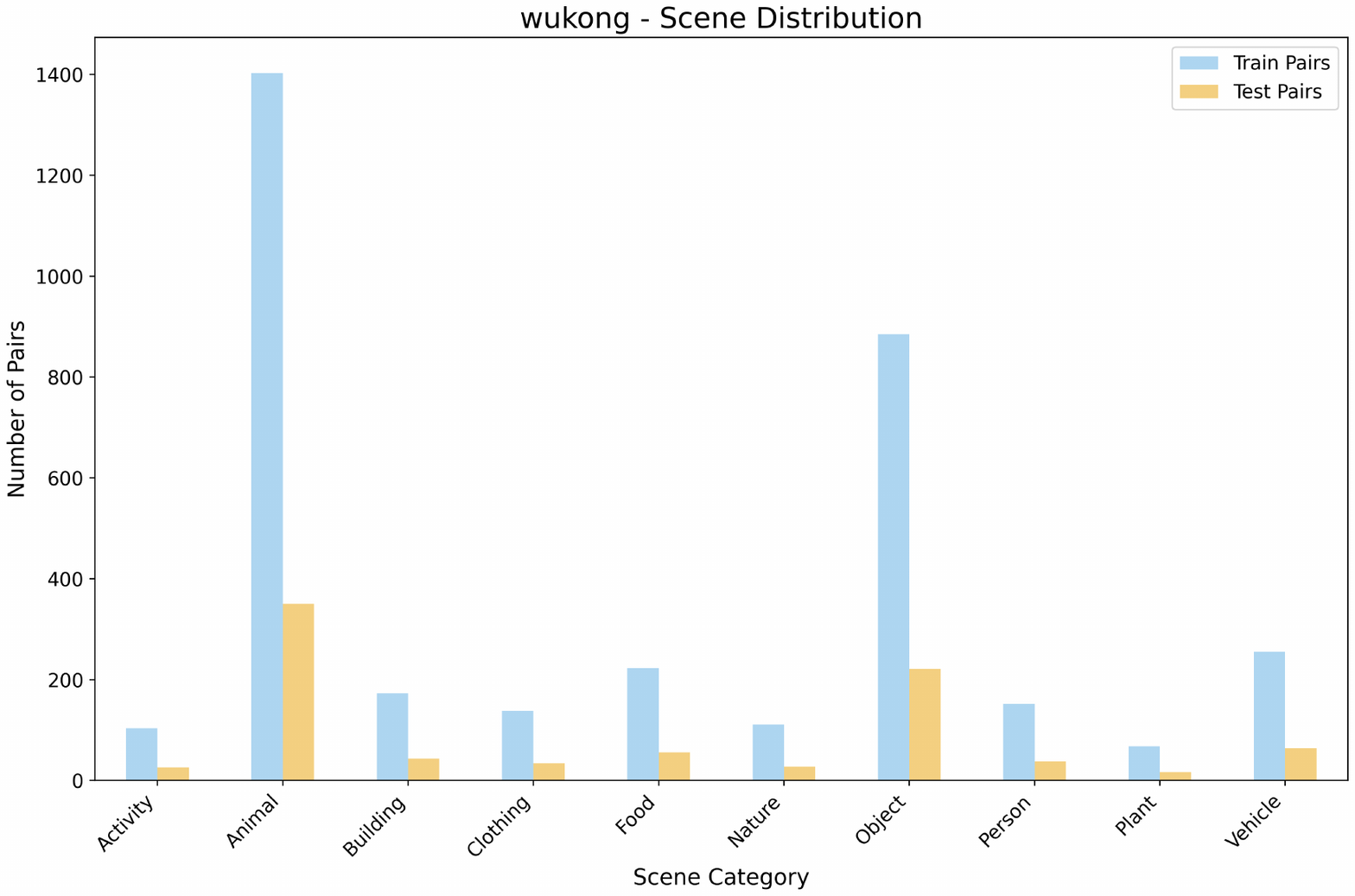} 
    \caption{The scene distribution of Wukong}
    \label{Wukong} 
\end{figure}

\noindent
\textbf{Wukong \cite{wukong2022}.} A total of 3.5K sample pairs are randomly selected from the dataset provided by GenImage \cite{genimage} for the training set. Real images are selected from the ImageNet dataset, while generative models are trained using ImageNet. The scene distribution in the dataset is shown in \cref{Wukong}.

\begin{figure}[H] 
    \centering 
    \includegraphics[width=\linewidth]{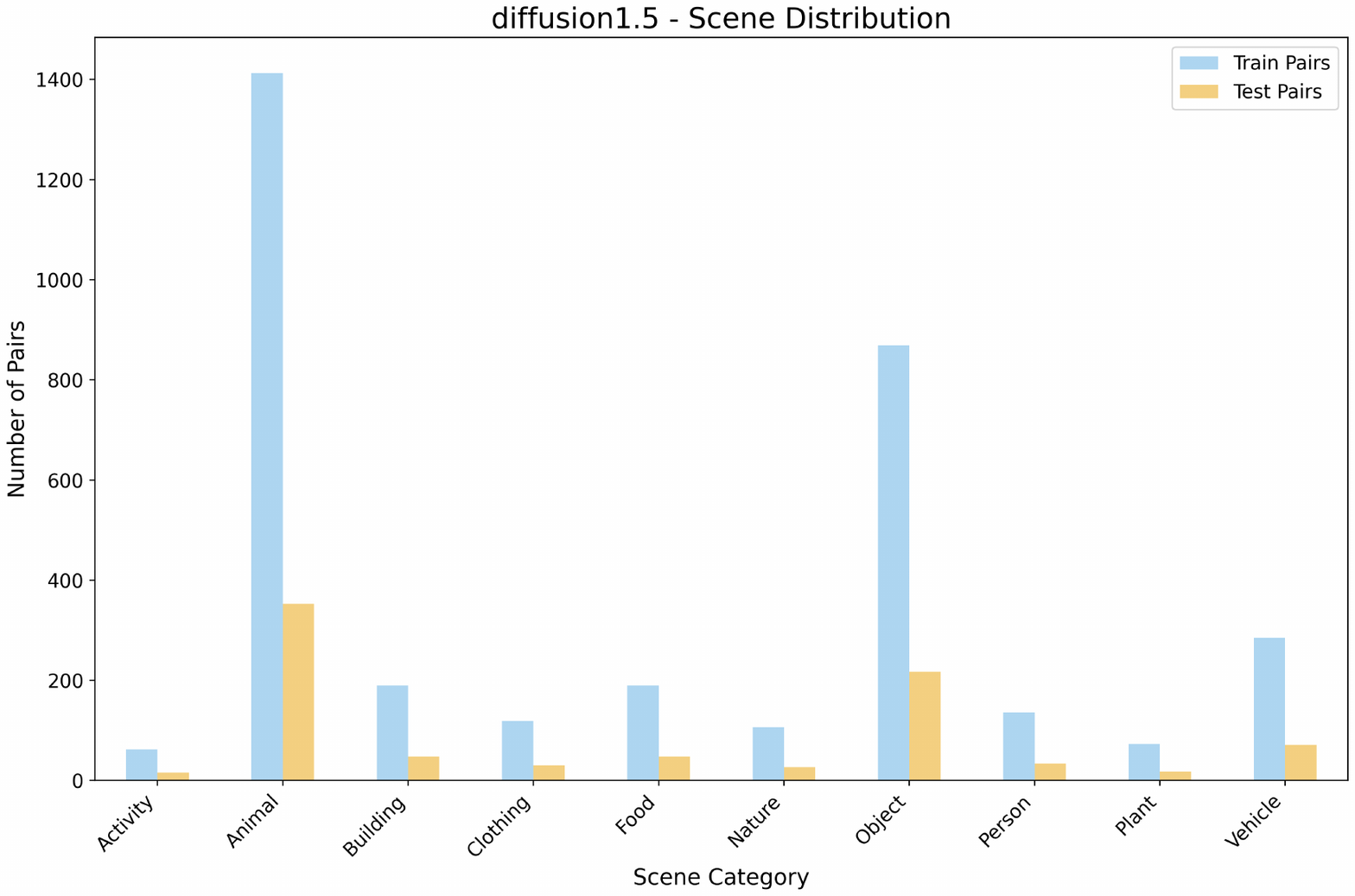} 
    \caption{The scene distribution of SD1.5}
    \label{SD1.5} 
\end{figure}

\noindent
\textbf{SD1.5 \cite{SD1.5}.} A total of 3.5K sample pairs are randomly selected from the dataset provided by GenImage \cite{genimage} for the training set. Real images are selected from the ImageNet dataset, while generative models are trained using ImageNet. The scene distribution in the dataset is shown in \cref{SD1.5}.

\begin{figure}[H] 
    \centering 
    \includegraphics[width=\linewidth]{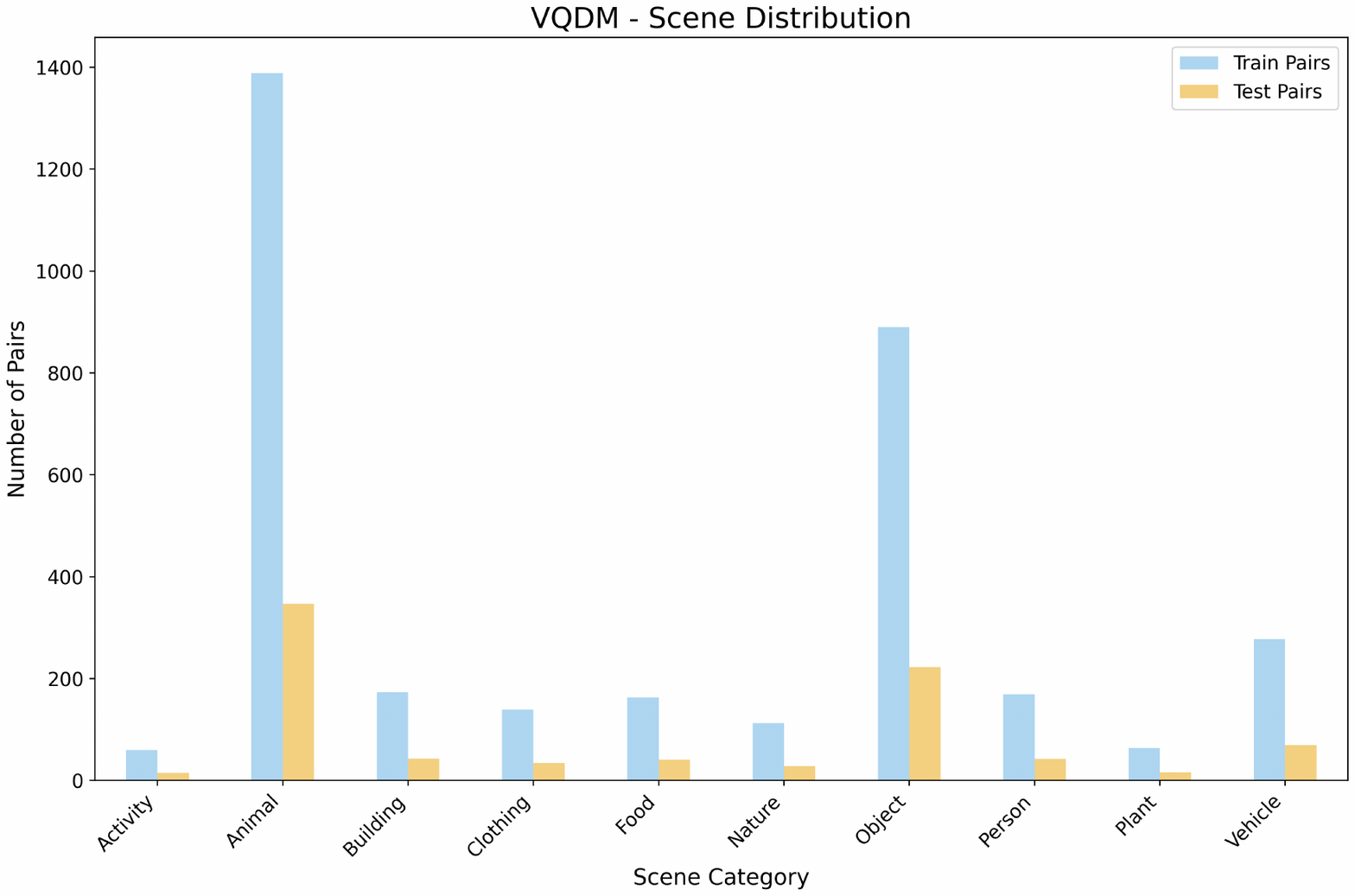} 
    \caption{The scene distribution of VQDM}
    \label{VQDM} 
\end{figure}

\noindent
\textbf{VQDM \cite{VQDM}.} A total of 3.5K sample pairs are randomly selected from the dataset provided by GenImage \cite{genimage} for the training set. Real images are provided by the ImageNet dataset, while generative models are trained using ImageNet. The scene distribution in the dataset is shown in \cref{VQDM}.

\begin{figure}[H] 
    \centering 
    \includegraphics[width=\linewidth]{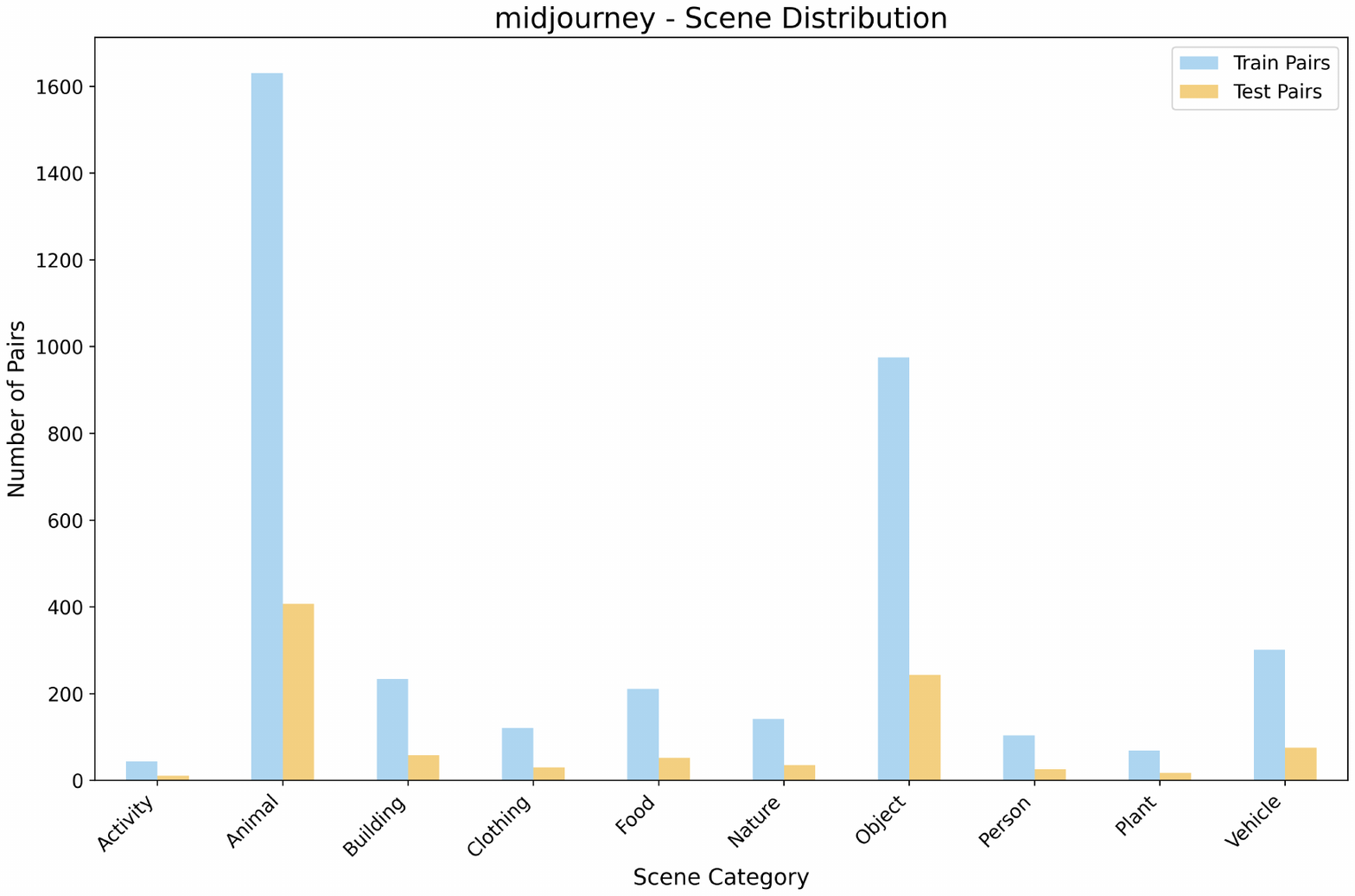} 
    \caption{The scene distribution of Midjourney-V5}
    \label{Midjourney-V5} 
\end{figure}

\noindent
\textbf{Midjourney-V5 \cite{midjourney2022}.} A total of 3.5K sample pairs are randomly selected from the dataset provided by GenImage \cite{genimage} for the training set. Real images are selected from the ImageNet dataset, while generative models are trained using ImageNet. The scene distribution in the dataset is shown in \cref{Midjourney-V5}.

\subsection{Open-World Dataset}
To evaluate the generalization ability of different methods when facing open-world datasets, this section considers generative models that are more advanced than those in the continual learning dataset. This setup also reflects the models’ adaptability when encountering more sophisticated generative models.

\noindent
\textbf{StyleGAN-xl \cite{stylegan-xl}.} We collect 3K sample pairs as the open-world test set. To ensure rigorous evaluation of generalization, the distribution of this test set is kept as distinct as possible from that of the continual learning tasks. Therefore, real images are randomly selected from Open Images V7, while fake images are produced by generative models pretrained on ImageNet, as provided by StyleGAN-xl \cite{stylegan-xl}.

\begin{figure}[H] 
    \centering 
    \includegraphics[width=\linewidth]{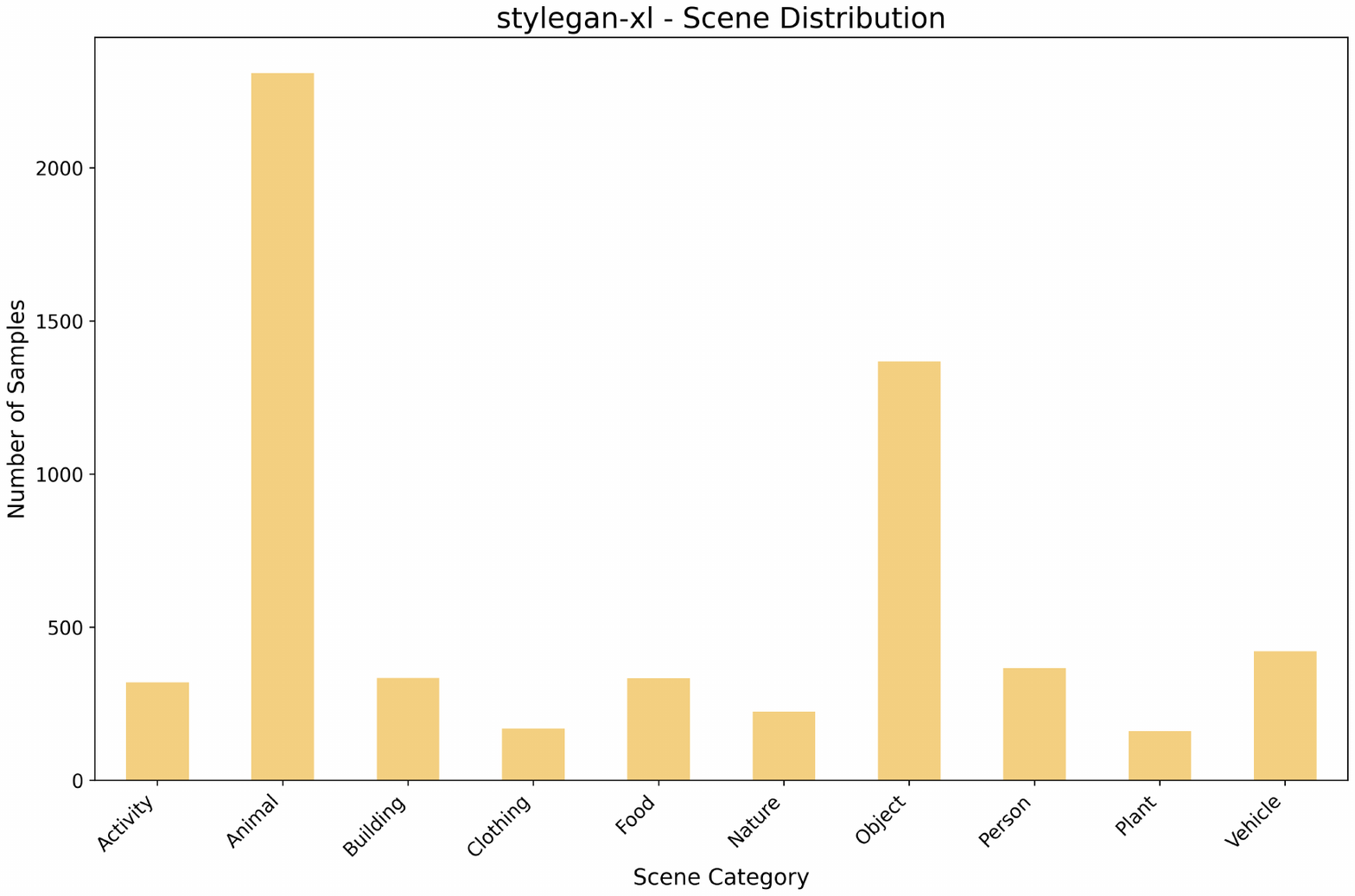} 
    \caption{The scene distribution of StyleGAN-xl}
    \label{StyleGAN-xl} 
\end{figure}

\noindent
\textbf{R3GAN \cite{R3GAN}.} A total of 3K sample pairs are used, with real images randomly drawn from Open Images V7 and fake images generated by ImageNet-pretrained models provided by R3GAN \cite{R3GAN}.

\begin{figure}[H] 
    \centering 
    \includegraphics[width=\linewidth]{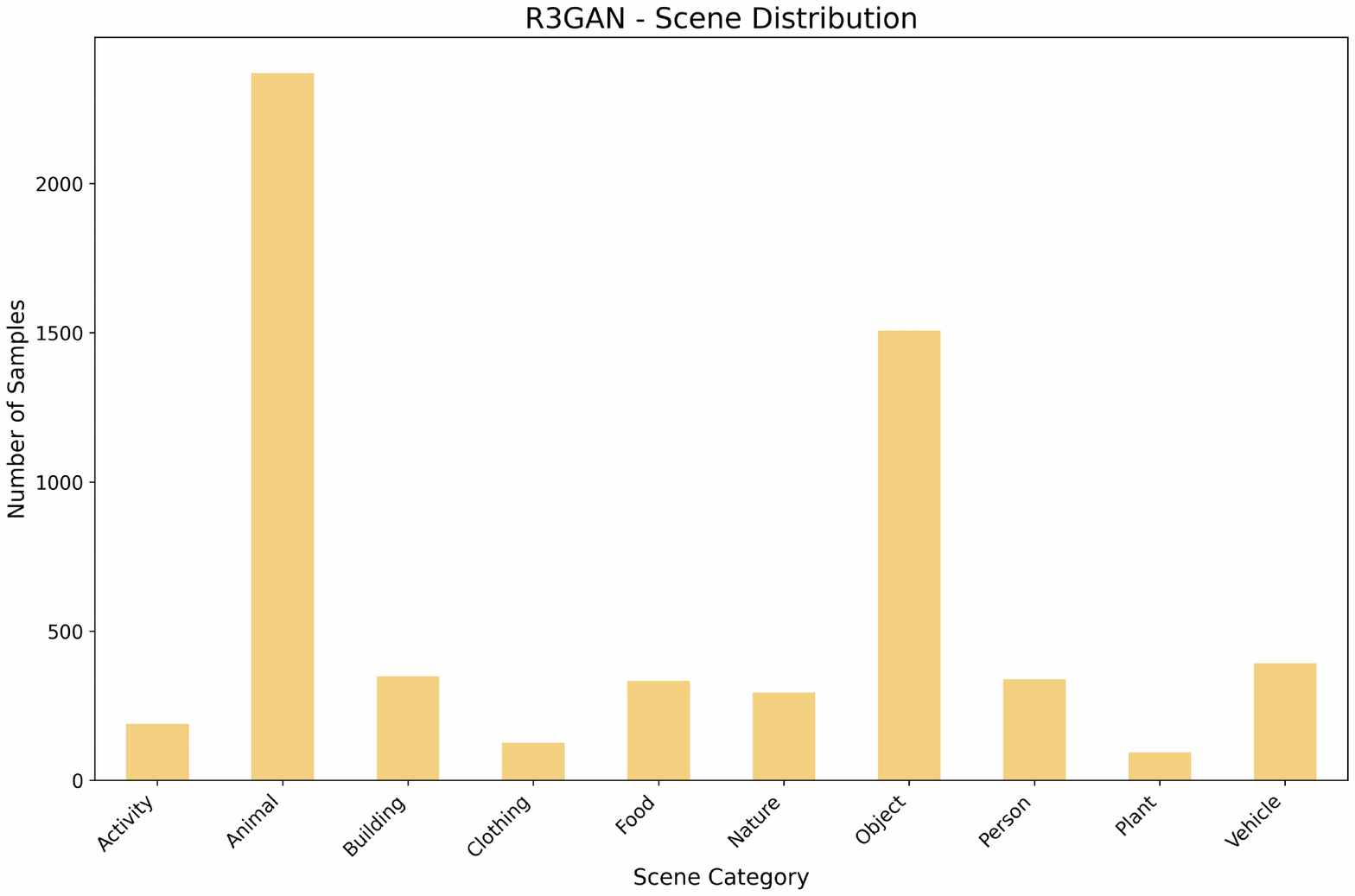} 
    \caption{The scene distribution of R3GAN}
    \label{R3GAN} 
\end{figure}

\noindent
\textbf{FLUX1-dev \cite{flux}.} A total of 4.5K sample pairs are randomly selected from AIGIBench \cite{AIGIBench}. Following AIGIBench settings, real images are drawn from Open Images V7, while fake images are generated using the FLUX1-dev pretrained diffusion model, which incorporates high-quality text prompts via the Gemini API.

\begin{figure}[H] 
    \centering 
    \includegraphics[width=\linewidth]{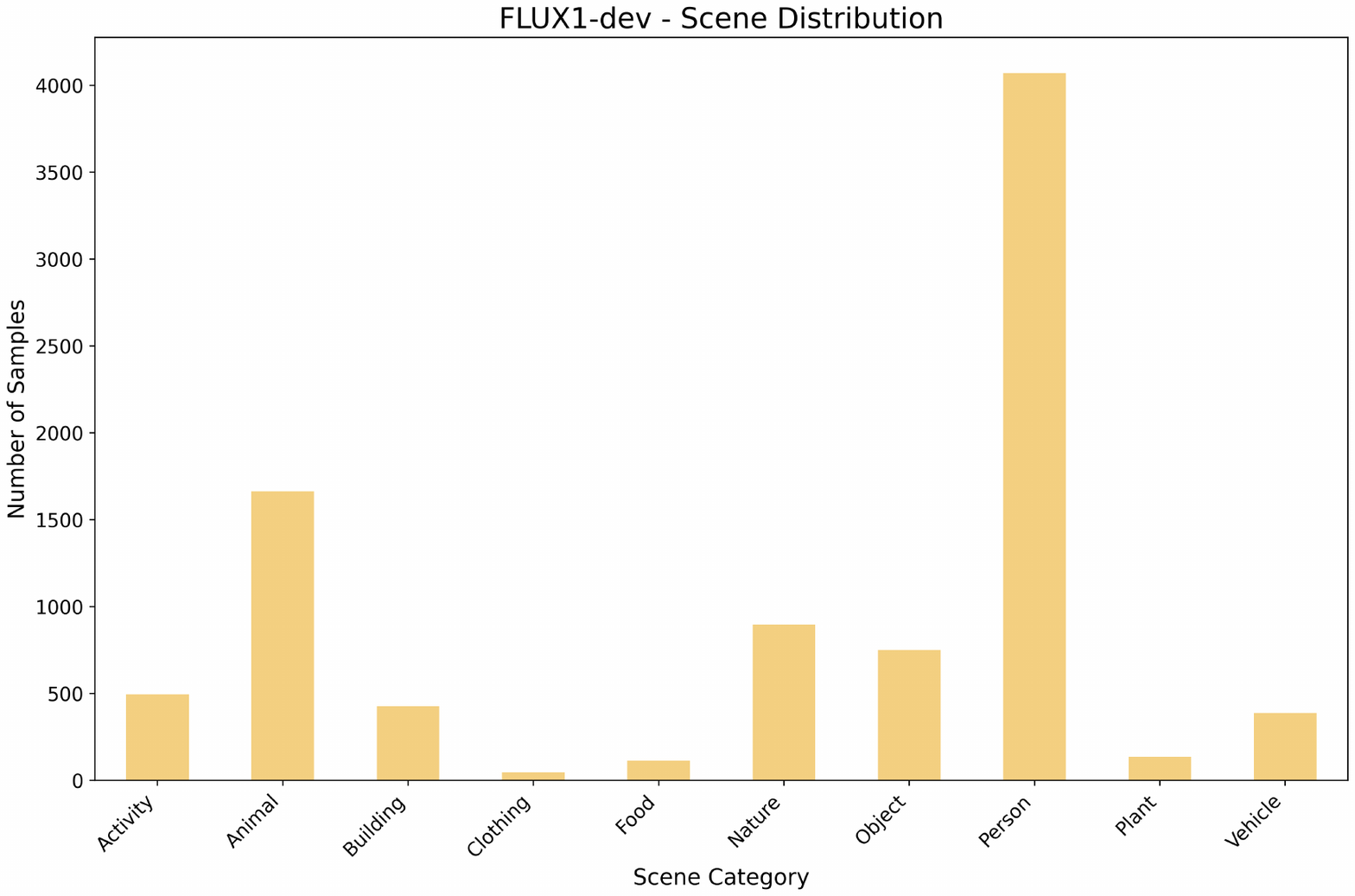} 
    \caption{The scene distribution of FLUX1-dev}
    \label{FLUX1-dev} 
\end{figure}

\noindent
\textbf{Midjourney-V6 \cite{midjourneyv6}.} A total of 3K sample pairs are randomly selected from AIGIBench \cite{AIGIBench}, with real images from Open Images V7 and fake images generated by the Midjourney-V6 pretrained diffusion model.

\begin{figure}[H] 
    \centering 
    \includegraphics[width=\linewidth]{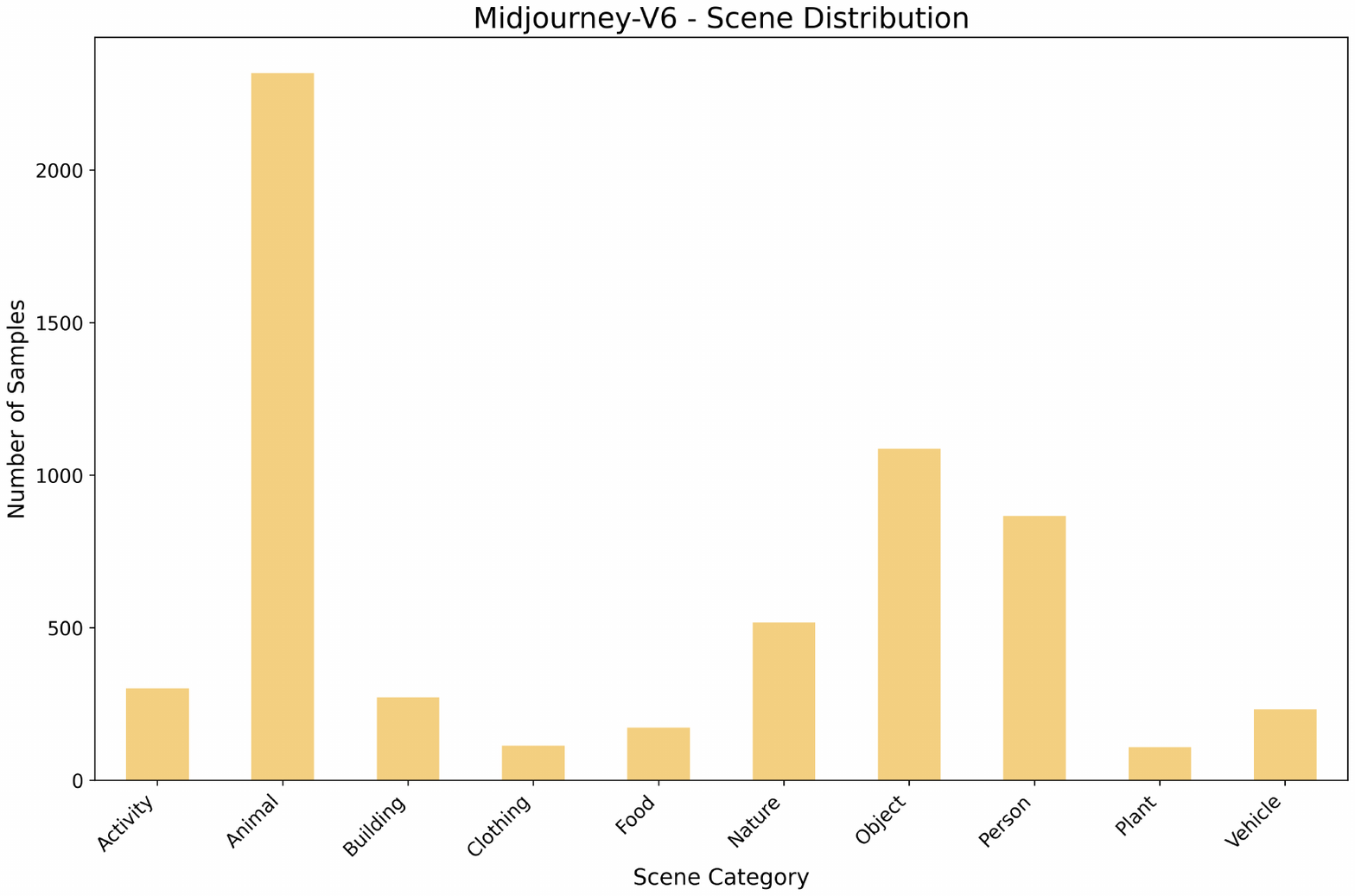} 
    \caption{The scene distribution of Midjourney-V6}
    \label{Midjourney-V6} 
\end{figure}

\noindent
\textbf{SD3 \cite{SD3}.} From AIGIBench \cite{AIGIBench}, 4.5K sample pairs are randomly selected, where real images come from Open Images V7 and fake images are produced by the Midjourney-V6 pretrained diffusion model.

\begin{figure}[H] 
    \centering 
    \includegraphics[width=\linewidth]{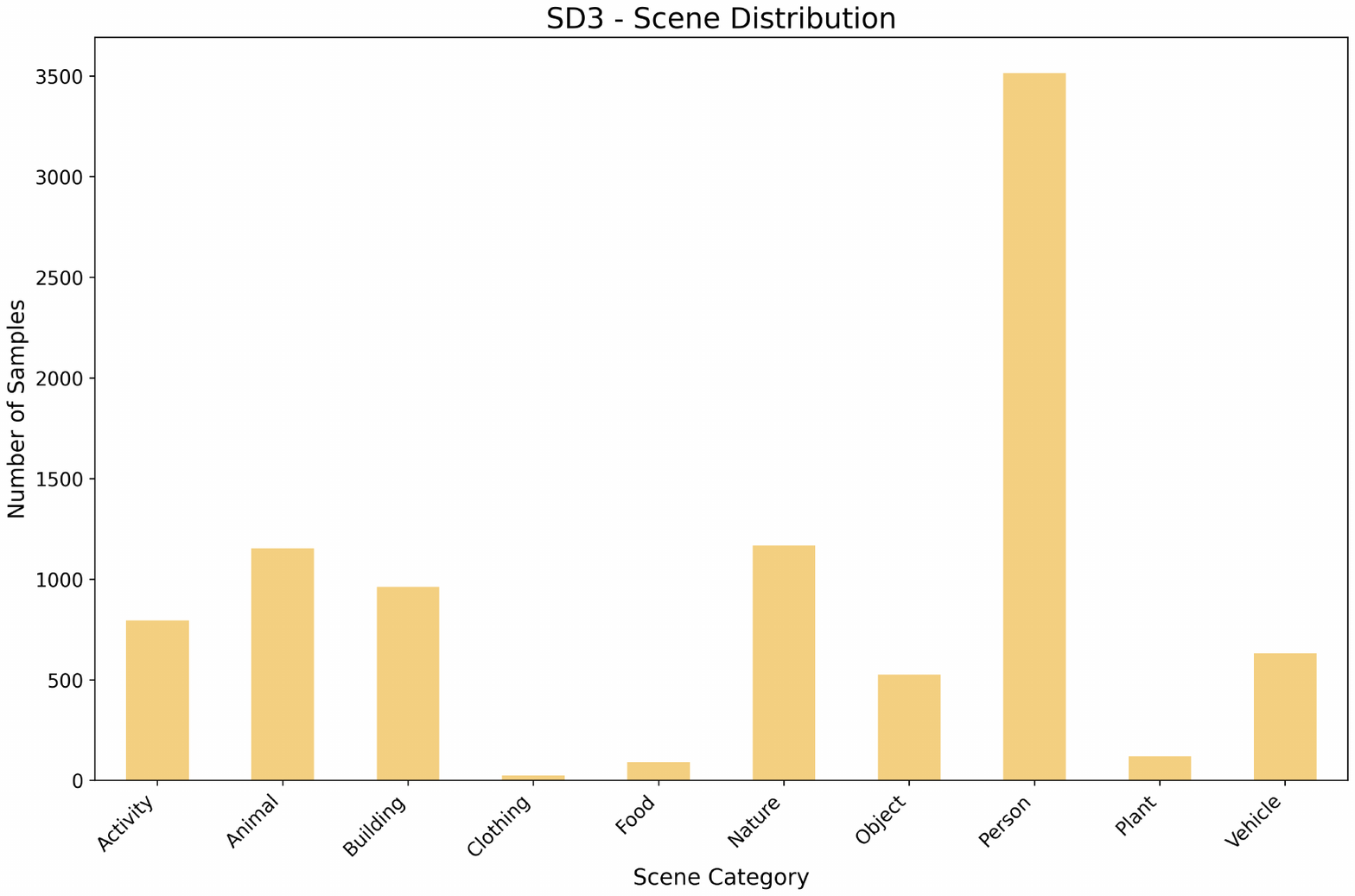} 
    \caption{The scene distribution of SD3}
    \label{SD3} 
\end{figure}

\noindent
\textbf{Imagen3 \cite{imagen3}.} A total of 4.5K sample pairs are randomly drawn from AIGIBench \cite{AIGIBench}, with real images sourced from Open Images V7 and fake images generated by the pretrained Midjourney-V6 diffusion model.

\begin{figure}[H] 
    \centering 
    \includegraphics[width=\linewidth]{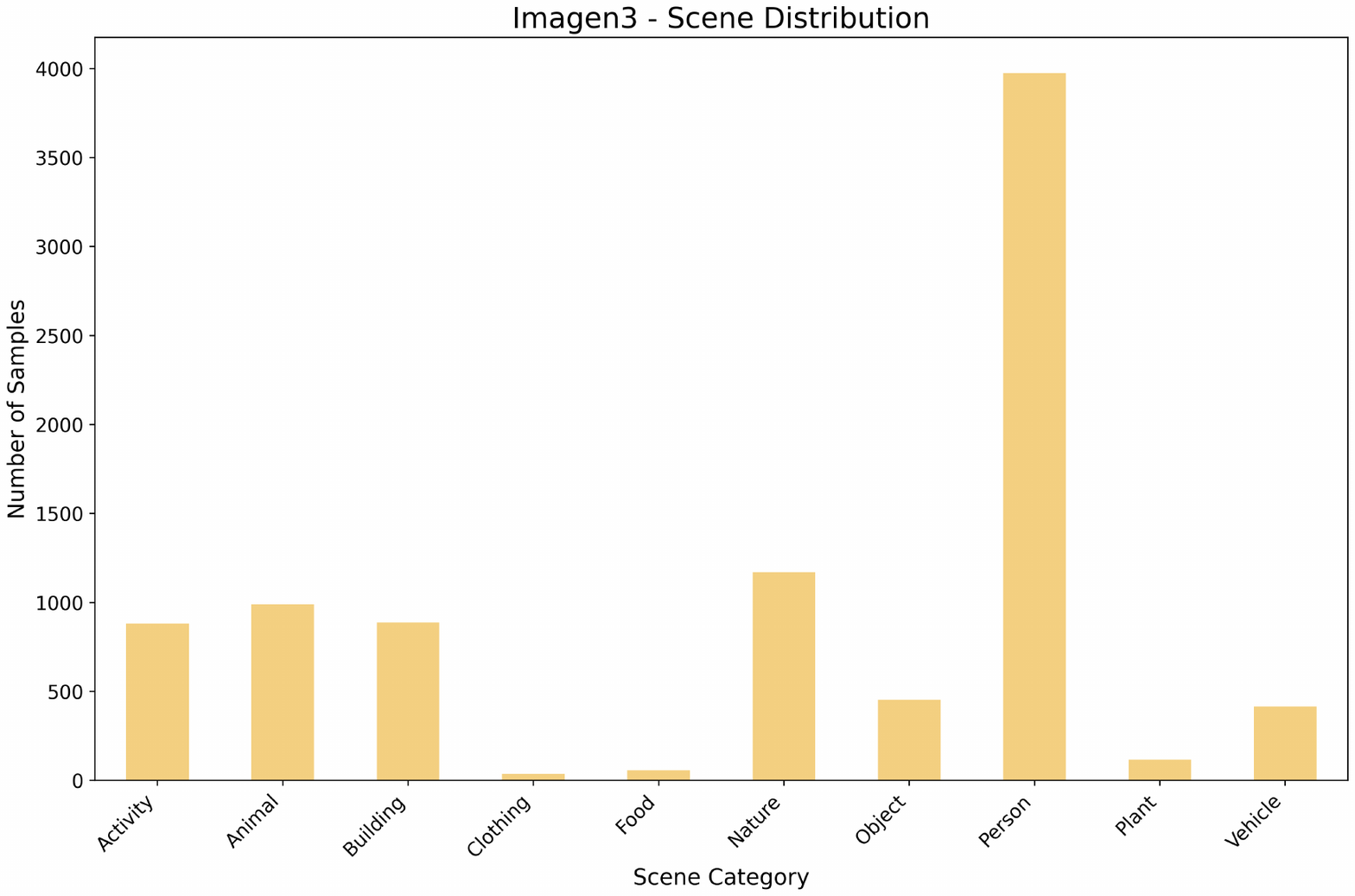} 
    \caption{The scene distribution of Imagen3}
    \label{Imagen3} 
\end{figure}

\section{Additional Order Configurations}

\begin{table*}[t]
    \centering
    \footnotesize
    \caption{Performance evaluation and comparison with other methods (\%), with the best results highlighted in bold, the second-best results underlined, and the results arranged according to Extra Order 1.}
    \label{tab:performance_comparison_in_continual_learning_1}

\begin{subtable}{\textwidth}
\centering

\renewcommand{\arraystretch}{0.9}
\resizebox{\textwidth}{!}{
\begin{tabular}{c c c c *{6}{cc}}
    \toprule
    \multirow{2}{*}{Method} & \multirow{2}{*}{Venue} & \multirow{2}{*}{\shortstack{Continual\\Learning}} & \multirow{2}{*}{\shortstack{Replay\\Set}} & \multicolumn{1}{c}{1-ADM} & \multicolumn{2}{c}{2-GLIDE} & \multicolumn{2}{c}{3-SAGAN} & \multicolumn{2}{c}{4-ProGAN} \\
    \cmidrule(lr){5-11}
    & & & & AA & AA & AF & AA & AF & AA & AF \\
    \midrule
    CLIP+LoRA & - & $\times$ & $-$ & \textbf{99.99} & 96.87 & 6.26 & 92.78 & 10.83 & \underline{97.79} & \underline{2.09} \\
    \midrule
    Universe & CVPR'23 & $\times$ & $-$ & 93.70 & 76.27 & 19.53 & 89.80 & 8.79 & 85.04 & 17.38 \\
    NPR & CVPR'24 & $\times$ & $-$ & 92.86 & 92.56 & \textbf{-1.76} & 80.40 & 19.76 & 72.89 & 23.55 \\
    \midrule
    RegO & AAAI'25 & $\checkmark$ & $\times$ & 99.49 & 98.58 & 1.16 & 94.86 & 6.77 & \textbf{97.84} & \textbf{1.10} \\
    Tang et al. & TIFS'25 & $\checkmark$ & $\checkmark$ & 99.35 & \underline{99.06} & \underline{0.07} & \underline{97.51} & \textbf{1.18} & 95.09 & 2.82 \\
    \rowcolor{cyan!10}
    SAIDO (Ours) & - & $\checkmark$ & $\times$ & \underline{99.81} & \textbf{99.61} & 0.32 & \textbf{98.40} & \underline{2.17} & 97.03 & 2.90 \\
    \bottomrule
\end{tabular}}
\renewcommand{\arraystretch}{1}
\end{subtable}

\vspace{0.4em}

\begin{subtable}{\textwidth}
\centering
\resizebox{\textwidth}{!}{
\begin{tabular}{c *{10}{cc} c}
    \toprule
    \multirow{2}{*}{Method} & \multicolumn{2}{c}{5-BigGAN} & \multicolumn{2}{c}{6-Wukong} & \multicolumn{2}{c}{7-Midjourney-V5} & \multicolumn{2}{c}{8-SD1.5} & \multicolumn{2}{c}{9-VQDM} & \multirow{2}{*}{\shortstack{New.\\ACC ($\uparrow$)}} \\
    \cmidrule(lr){2-11}
    & AA & AF & AA & AF & AA & AF & AA & AF & AA & AF & \\
    \midrule
    CLIP+LoRA & 94.34 & 6.43 & 76.46 & 27.57 & 84.63 & 17.08 & 82.62 & 19.02 & 71.46 & 31.37 & \textbf{99.34} \\
    \midrule
    Universe & 70.60 & 27.20 & 77.36 & 21.76 & 79.09 & 20.60 & 61.01 & 28.69 & 78.24 & 24.15 & 89.65 \\
    NPR  & 75.48 & 23.15 & 67.85 & 31.49 & 72.89 & 73.64 & 75.30 & 21.24 & 66.40 & 31.53 & 99.12 \\
    \midrule
    RegO & \underline{96.75} & 2.65 & 92.82 & 6.27 & 88.31 & 10.25 & 89.25 & 8.75 & 89.92 & 7.63 & 96.70 \\
    Tang et al. & 94.93 & \underline{2.48} & \underline{94.45} & \underline{2.98} & \underline{92.86} & \textbf{4.62} &	\underline{92.32} & \underline{5.09} & \underline{92.84} & \underline{4.61} & 97.91 \\
    \rowcolor{cyan!10}
    SAIDO (Ours) & \textbf{98.45} & \textbf{1.12} & \textbf{98.00} & \textbf{1.52} & \textbf{94.67} & \underline{5.16} & \textbf{96.38} & \textbf{3.16} & \textbf{96.22} & \textbf{3.27} & \underline{99.13} \\
    \bottomrule
\end{tabular}}
\end{subtable}
\end{table*}


\begin{table*}[t]
    \centering
    \footnotesize
    \caption{Performance evaluation and comparison with other methods under (\%), with the best results highlighted in bold, the second-best results underlined, and the results arranged according to Extra Order 2.}
    \label{tab:performance_comparison_in_continual_learning_2}

\begin{subtable}{\textwidth}
\centering

\renewcommand{\arraystretch}{0.9}
\resizebox{\textwidth}{!}{
\begin{tabular}{c c c c *{6}{cc}}
    \toprule
    \multirow{2}{*}{Method} & \multirow{2}{*}{Venue} & \multirow{2}{*}{\shortstack{Continual\\Learning}} & \multirow{2}{*}{\shortstack{Replay\\Set}} & \multicolumn{1}{c}{1-ADM} & \multicolumn{2}{c}{2-ProGAN} & \multicolumn{2}{c}{3-Wukong} & \multicolumn{2}{c}{4-SAGAN} \\
    \cmidrule(lr){5-11}
    & & & & AA & AA & AF & AA & AF & AA & AF \\
    \midrule
    CLIP+LoRA & - & $\times$ & $-$ & \textbf{99.86} & \underline{98.09} & 1.21 & 85.69 & 19.66 & 85.43 & 18.21 \\
    \midrule
    Universe & CVPR'23 & $\times$ & $-$ & 93.70 & 90.06 & 17.95 & 83.17 & 30.19 & 81.98 & 30.1 \\
    NPR & CVPR'24 & $\times$ & $-$ & 92.86 & 86.18 & 14.10 & 72.89 & 29.06 & 63.24 & 40.37 \\
    \midrule
    RegO & AAAI'25 & $\checkmark$ & $\times$ & 99.49 & 97.91 & \underline{0.65} & 90.76 & 8.74 & 92.92 & \underline{5.77} \\
    Tang et al. & TIFS'25 & $\checkmark$ & $\checkmark$ & 99.21 & 96.76 & 1.76 & \underline{93.73} & \underline{5.48} & \underline{90.71} & 9.34 \\
    \rowcolor{cyan!10}
    SAIDO (Ours) & - & $\checkmark$ & $\times$ & \underline{99.69} & \textbf{98.27} & \textbf{0.56} & \textbf{97.74} & \textbf{1.48} & \textbf{96.84} & \textbf{2.79} \\
    \bottomrule
\end{tabular}}
\renewcommand{\arraystretch}{1}
\end{subtable}

\vspace{0.4em}

\begin{subtable}{\textwidth}
\centering
\resizebox{\textwidth}{!}{
\begin{tabular}{c *{10}{cc} c}
    \toprule
    \multirow{2}{*}{Method} & \multicolumn{2}{c}{5-BigGAN} & \multicolumn{2}{c}{6-SD1.5} & \multicolumn{2}{c}{7-GLIDE} & \multicolumn{2}{c}{8-VQDM} & \multicolumn{2}{c}{9-Midjourney-V5} & \multirow{2}{*}{\shortstack{New.\\ACC ($\uparrow$)}} \\
    \cmidrule(lr){2-11}
    & AA & AF & AA & AF & AA & AF & AA & AF & AA & AF & \\
    \midrule
    CLIP+LoRA & 87.05 & 15.26 & \underline{94.10} & 6.26 & 89.73 & 11.29 & \underline{97.53} & \underline{2.17} & 90.32 & 10.16 & \textbf{99.36} \\
    \midrule
    Universe & 85.93 & 26.03 & 81.85 & 30.81 & 54.41 & 33.82 & 80.54 & 35.58 & 77.93 & 30.15 & 90.68 \\
    NPR & 69.05 & 31.65 & 80.43 & 17.42 & 62.21 & 35.78 & 64.66  & 33.08 & 72.46 & 23.12 & 93.01 \\
    \midrule
    RegO  & \underline{94.25} & \underline{4.10} & 92.90 & 4.97 & \underline{92.89} & \underline{5.04} & 92.68 & 5.05 & 88.19 & 9.30 & 96.45 \\
    Tang et al. & 90.17 & 9.84 & 94.08 & \underline{4.81} & 91.79 & 7.56 & 93.14 & 5.70 & \underline{92.58} & \underline{5.92} & 97.84 \\
    \rowcolor{cyan!10}
    SAIDO (Ours) & \textbf{96.75} & \textbf{2.94} & \textbf{98.25} & \textbf{1.09} & \textbf{97.47} & \textbf{2.22} & \textbf{98.27} & \textbf{1.16} & \textbf{94.26} & \textbf{5.55} & \underline{99.19} \\
    \bottomrule
\end{tabular}}
\end{subtable}
\end{table*}

To ensure comprehensive experimental settings, nine generative image models are considered in the continual learning tasks, and long-sequence continual learning experiments with multiple task orders are conducted. This section presents and analyzes the performance comparisons for the alternative task sequences.

\begin{itemize}
    \item \textbf{Extra Order 1.} $\mathcal{D}_1$ = \{ADM, GLIDE, SAGAN, ProGAN, BigGAN, Wukong, Midjourney-V5, SD1.5, VQDM\}.
    \item \textbf{Extra Order 2.} $\mathcal{D}_2$ = \{ADM, ProGAN, Wukong, SAGAN, BigGAN, SD1.5, GLIDE, VQDM, Midjourney-V5\}.
\end{itemize}

As shown in \cref{tab:performance_comparison_in_continual_learning_1} and \cref{tab:performance_comparison_in_continual_learning_2}, our method achieves superior performance compared with SOTA methods under various task order settings. This indicates that our method is less susceptible to variations in the order of continual learning tasks, demonstrating its stability. Notably, our method exhibits substantially less degradation in overall performance across various task sequences compared to current SOTA methods. Under the task order of Protocol 1, our method not only achieves the highest average accuracy on most tasks, but also maintains an average accuracy of 99.81\% on the first task (ADM) and 95.61\% on the final task (Midjourney-V5). In contrast, the current SOTA method drops from 99.21\% to 92.13\% over the same progression. Similar trends are observed under other task orders, highlighting the superior balance between plasticity and stability provided by our method.

\end{document}